\pdfoutput=1
\documentclass[10pt,twocolumn,letterpaper]{article}

\usepackage{cvpr}              % To produce the CAMERA-READY version
\usepackage{graphicx}
\usepackage{amsmath}
\usepackage{amssymb}
\usepackage{booktabs}
\usepackage[accsupp]{axessibility}
\usepackage{mathtools}
\usepackage{amsmath}

\usepackage{amssymb}
\usepackage{amsfonts}
\usepackage{amsopn}
\usepackage{graphicx} % Necessary to use \scalebox
\usepackage{textcomp}
\usepackage{xfrac}
\usepackage{bbm}
\usepackage{overpic}
\usepackage[ruled,vlined,linesnumbered]{algorithm2e}
\SetAlFnt{\small}
\SetAlCapFnt{\small}
\SetAlCapNameFnt{\small}
\SetAlCapHSkip{0pt}

\usepackage{multirow}

\usepackage{booktabs}
\usepackage{footnote}
\usepackage{paralist}
\usepackage{autobreak}
\usepackage{wrapfig}
\usepackage{color}
\definecolor{green}{rgb}{0, 0.6, 0}
\definecolor{orange}{rgb}{0.8, 0.6, 0.2}
\definecolor{red}{rgb}{1.0, 0.0, 0.0}
\definecolor{teal}{rgb}{0.0, 0.4, 0.4}
\definecolor{purple}{rgb}{0.65,0,0.65}
\definecolor{saffron}{rgb}{0.95,0.75,0.2}
\definecolor{turquoise}{rgb}{0.0,0.5,0.5}
\definecolor{brown}{rgb}{0.5, 0.16, 0.16}

\usepackage{overpic}
\usepackage{currfile} % \currfiledir

\newlength\savedwidth

\definecolor{lightgray}{rgb}{0.6, 0.6, 0.6}

\usepackage[normalem]{ulem}

\newcommand{\hidecomment}[1]{}
\newcommand{\R}{\mathbb{R}}
\newcommand{\M}{\mathcal{M}}
\newcommand{\B}{\mathcal{B}}

\newcommand{\A}{\mathcal{A}}

\newcommand{\N}{\mathcal{N}}

\newcommand{\bO}{\mathcal{O}}

\usepackage{xspace}
\usepackage{url}

\usepackage{soul}

\usepackage{xcolor}

\newcommand{\de}[1]{{\color{gray}}}

\usepackage[pagebackref,breaklinks,colorlinks]{hyperref}

\usepackage[capitalize]{cleveref}
\crefname{section}{Sec.}{Secs.}
\Crefname{section}{Section}{Sections}
\Crefname{table}{Table}{Tables}
\crefname{table}{Tab.}{Tabs.}

\begin{document}

%\title{\refine{Do One Thing but Think Twice (or Think Twice before You Act): Simultaneous (\liudai{or Parallel?}) Exploration and Verification for 3D-Aware Object Goal Navigation}}
\title{3D-Aware Object Goal Navigation\\ via Simultaneous Exploration and Identification}

\author{Jiazhao Zhang$^{1,2,\ast}$ \quad\quad Liu Dai$^{3,}$\thanks{Joint first authors}
\quad\quad Fanpeng Meng$^{4}$ \quad\quad Qingnan Fan$^{5}$ \\ \quad\quad Xuelin Chen$^{5}$ \quad\quad Kai Xu$^{6}$ \quad\quad He Wang$^{1}$\thanks{Corresponding author: hewang@pku.edu.cn}\\
\normalsize{ $^1$ CFCS, Peking University \quad $^2$ Beijing Academy of Artificial Intelligence \quad $^3$CEIE, Tongji University}  \\ \normalsize{ \quad $^4$Huazhong University of Science and Technology \quad $^5$Tencent AI Lab \quad $^6$ National University of Defense Technology} 
}

% {$^1$ Peking University \quad $^2$ BAAI \quad $^3$CEIE, Tongji University  \quad $^4$HUST \quad $^5$Tencent AI Lab \quad $^6$ NUDT 

% Huazhong University of Science and Technology
% National University of Defense Technology

% \and
\maketitle
%%%%%%%%% ABSTRACT
\begin{abstract}
% Object goal navigation (ObjectNav) in unseen environments is a fundamental task for Embodied AI. Agents in existing works learn their ObjectNav policies based on 2D maps, scene graphs, or image sequences. Considering this task happens in 3D space, a 3D-aware agent can advance the ObjectNav capability via learning from fine-grained spatial information. However, leveraging 3D scene representation can be prohibitively unpractical for policy learning in this floor-level task, due to low sampling efficiency and expensive computational costs. In this work, we enable the 3D-aware ObjectNav through learning two sub-polices, namely corner-guided exploration policy and category-aware verification policy, both utilizing online fused 3D points as observation. Through extensive experiments, we find that this framework can dramatically improve the performance in ObjectNav through learning 3D scene representation. Our framework achieves the best performance among all modular-based methods on the Matterport3D and Gibson datasets, while requiring (up to 30x) less computational cost for training.

Object goal navigation (ObjectNav) in unseen environments is a fundamental task for Embodied AI.
%
% XL: what is problematic with exisiting works: they are not explicitly 3D-aware? by explicitly, I mean a scene graph may also be a graph resides in 3D?
Agents in existing works learn ObjectNav policies based on 2D maps, scene graphs, or image sequences. 
% \Considering object goal nanvgiation happens in 3D space, 
% XL: why 3D-aware, motivation and advantages %forming 3D scene representation
% Considering an established fact that target objects lie in 3D space,
Considering this task happens in 3D space, a 3D-aware agent can advance its ObjectNav capability via learning from fine-grained spatial information.
% \refine{\sout{Despite the advantages of the 3D-aware agent} 
% XL: why previous methods not 3D-aware, what is the challenge
% However, learning ObjectNav policy from 3D representation can be prohibitively unpractical  tasks because of the low sampling efficiency and expensive computational costs.
% However, floor-level 3D scene representation \liudai{for the task} can be hardly \sout{applied to}
%  \liudai{leveraged in}\sout{ObjectNav} policy learning due to low sampling efficiency and expensive computational costs.
However, leveraging 3D scene representation can be prohibitively unpractical for policy learning in this floor-level task, due to low sample efficiency and expensive computational cost.
%learning a Object nav policy from 3D ... is prohibitively unpractical ...
% XL: how we address the challenge - a high-level description
In this work, 
% \liudai{we enable the 3D-aware ObjectNav through a decomposition framework, where the agent needs to learn two sub-polices, namely exploration policy and verification policy.}
we propose a framework for the challenging 3D-aware ObjectNav based on two straightforward sub-policies.
The two sub-polices, namely corner-guided exploration policy and category-aware identification policy, simultaneously perform by utilizing online fused 3D points as observation.
% \liudai{These two sub-tasks a re each tackled by a policy running simultaneously to guide the agent with an exploration goal or a verified object goal(if exist), both leveraging online fused 3D points as observation.} 
Through extensive experiments, we show that this framework can dramatically improve the performance in ObjectNav through learning from 3D scene representation.
Our framework achieves the best performance among all modular-based methods on the Matterport3D and Gibson datasets, while requiring (up to 30x) less computational cost for training. The code will be released to benefit the community.\footnote{Homepage: {https://pku-epic.github.io/3D-Aware-ObjectNav/}}

\end{abstract}

%%%%%%%%% BODY TEXT
\section{Introduction}
\label{sec:intro}
As a vital task for intelligent embodied agents,
object goal navigation (ObjectNav) \cite{savva2019habitat,habitatchallenge2022}
requires an agent to find an object of a particular category in an unseen and unmapped scene.
% Recently, 
% the community has witnessed an \warning{emerging progress} in ObjectNav, where researchers propose to learn scene priors over various scene representations, e.g., 2D maps~\cite{ramakrishnan2022poni,georgakis2022l2m,chaplot2020object}, scene graphs~\cite{zhu2021soon} or directly over RGBD sequences~\cite{ye2021auxiliary, ramrakhya2022habitat,maksymets2021thda}.
% %
% \refine{These representations are widely-studied, by performing as abstract alternatives of actual 3D scenes.}
% Existing works tackle this task through end-to-end RL or modular-based methods. End-to-end RL has made good progress by directly outputting low-level actions based on video sequences, while suffering from poor generalizability to novel scenes. scenes~\cite{ramakrishnan2022poni,ye2021auxiliary,}.
Existing works tackle this task through end-to-end reinforcement learning (RL)~\cite{wijmans2019dd,ye2021auxiliary,ramrakhya2022habitat,maksymets2021thda} or modular-based methods~\cite{chaplot2020object,georgakis2022l2m,ramakrishnan2022poni}. End-to-end RL based methods take as input the image sequences and directly output low-level navigation actions, achieving competitive performance while suffering from lower sample efficiency and poor generalizability across datasets~\cite{campari2020exploiting,maksymets2021thda}. Therefore, we favor modular-based methods, which usually contain the following modules: 
a semantic scene mapping module that aggregates the RGBD observations and the outputs from semantic segmentation networks to form a semantic scene map;
an RL-based goal policy module that takes as input the semantic scene map and learns to online update a goal location; 
finally, a local path planning module that drives the agent to that goal.
Under this design, the semantic accuracy and geometric structure of the scene map are crucial to the success of object goal navigation.

\begin{figure}[t]
\centering
\begin{overpic}
[width=1.0\linewidth]{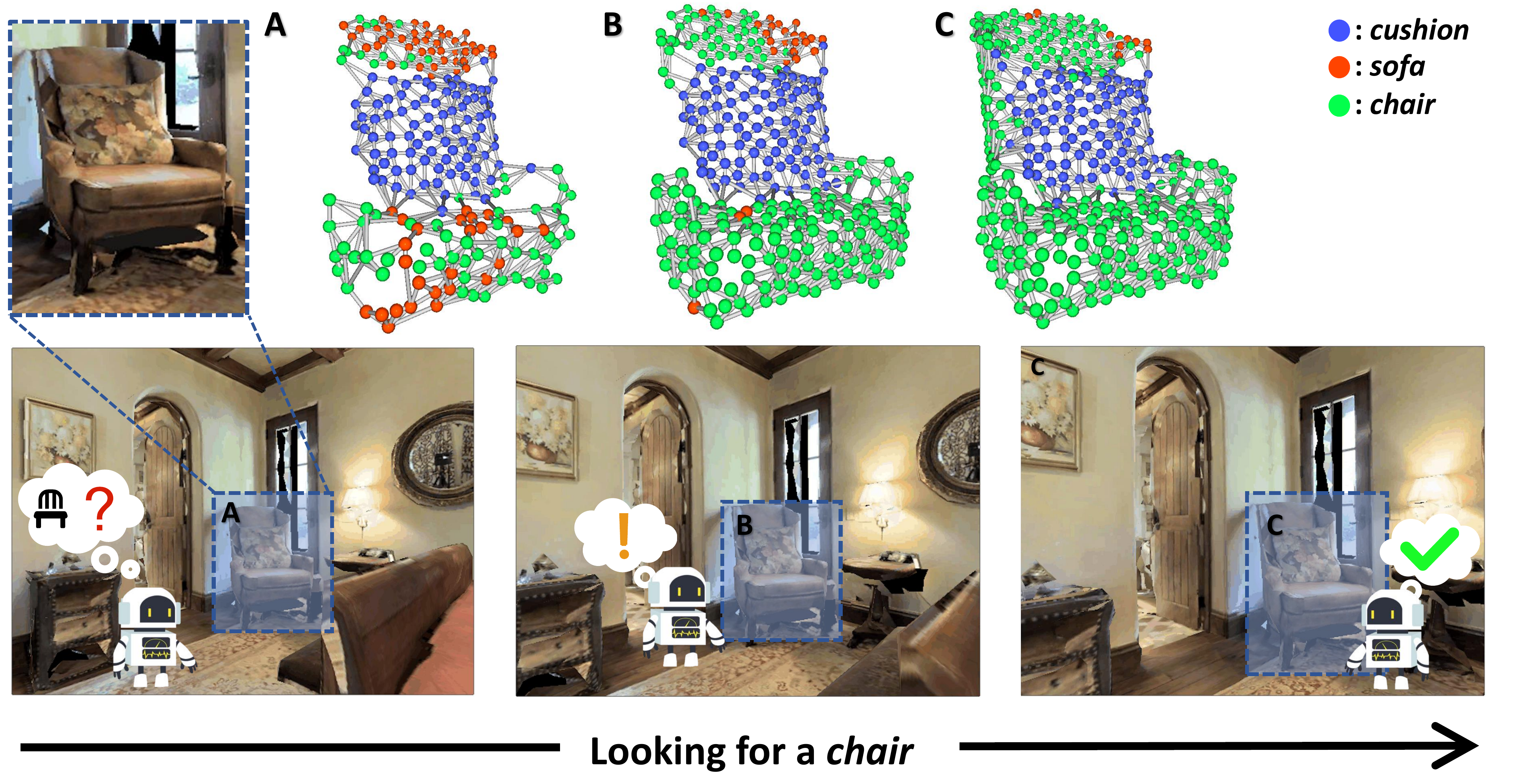}
\end{overpic}
\caption{We present a 3D-aware ObjectNav framework along with simultaneous exploration and identification policies: \textbf{A$\rightarrow$B}, the agent was guided by an exploration policy to look for its target; \textbf{B $\rightarrow$ C}, the agent consistently identified a target object and finally called STOP.}
\vspace{-0.8cm}
\label{fig:teaser}
\end{figure}

We observe that the existing modular-based methods mainly construct
2D maps~\cite{chaplot2020learning,chaplot2020object}, scene graphs~\cite{zhu2021soon,qiu2020learning} or neural fields~\cite{shafiullah2022clip} as their scene maps.
Given that objects lie in 3D space, these scene maps are inevitably deficient in leveraging 3D spatial information of the environment comprehensively and thus have been a bottleneck for further improving object goal navigation.
In contrast, forming a 3D scene representation naturally
offers more accurate, spatially dense and consistent semantic predictions than its 2D counterpart, as proved by ~\cite{Dai20183DMVJ3, Nekrasov2021Mix3DOD, Vu2022SoftGroupF3}.
Hence, if the agent could take advantage of the 3D scene understanding and form a 3D semantic scene map,
it is expected to advance the performance of ObjectNav. 

However, 
leveraging 3D scene representation would bring great challenges to ObjectNav policy learning.
% informative -> high-dimensional; use digest; during training; couple with Reinforcement learning
First, building and querying fine-grained 3D representation across a floor-level scene requires extensive computational cost, which can significantly slow down the training of RL~\cite{chaplot2021seal,Zheng2019ActiveSU}.
Also, 3D scene representation induces considerably more complex and high-dimensional observations to the goal policy than its 2D counterpart, leading to a lower sample efficiency and hampering the navigation policy learning~\cite{Zhu2017TargetdrivenVN, Lin2020Modeling3S}. 
As a result, it is demanding to design a framework to efficiently and effectively leverage powerful 3D information for ObjectNav.
% \textit{First,}
% \refine{building and querying fine-grained 3D scene representation requires extensive computational cost which leads to an order-of-magnitude running efficiency drop}~\cite{chaplot2021seal,Zheng2019ActiveSU};
% \textit{Second,}
% \refine{3D scene representation induces more complex and informative observation than its 2D alternatives. This make the agent require more \warning{episodes} which is crucial for reinforcement learning}
% ~\cite{Zhu2017TargetdrivenVN, Lin2020Modeling3S}.
% These two challenges collectively hinder the use of 3D representations in the ObjectNav task. 

% each policy is responding to task?  two goals -> task 

To tackle these challenges, we propose a novel framework composed of an online semantic point fusion module for 3D semantic scene mapping and two parallel policy networks in charge of scene exploration and object identification, along with a local path planning module.
Our online semantic point fusion module extends a highly efficient online
point construction algorithm \cite{zhang2020fusion} to enable online semantic fusion and spatial semantic consistency computation from captured RGBD sequences. This 3D scene construction empowers a comprehensive 3D scene understanding for ObjectNav. Moreover, compared to dense voxel-based methods\cite{chaplot2021seal,Zheng2019ActiveSU}, our point-based fusion algorithm are more memory-efficient\cite{whelan2015elasticfusion,schops2019bad} which makes it practically usable for floor-level navigation task. (See Figure~\ref{fig:teaser})

Moreover, to ease the learning of navigation policy, we further propose to factorize the navigation policy into two sub-policies, namely exploration and identification. The two policies simultaneously perform to roll out an exploration goal and an identified object goal (if exist), respectively. Then the input for the local path planning module will switch between these two goals, depending on whether there exists an identified target object. 
% \liudai{Here reveals the first part of the pun in our title where \textit{Think Twice} literally refers to that our agent would think about two policies during navigation: \textit{1) For now is there my target object?(verification policy)} \textit{2) Where should I go next if I haven't met my target 
% object?(exploration policy)}}
% corner goal *four*
More specifically, we propose a corner-guided exploration policy which learns to predict a long-term discrete goal at one of the four corners of the bounding box of the scene. These corner goals efficiently drive the agent to perceive the surroundings and explore regions where the target object is possibly settled.
% This design shares the same \sout{insights} \liudai{philosophy} of existing heuristic-based methods~\cite{luo2022stubborn} \sout{by avoiding back-and-forth paces} \liudai{which effectively avoids back-and-forth paces, while we endow the agent} with additional ability to encodes the scene's \liudai{semantic} priors.
And for identification, a category-aware identification policy is proposed to dynamically learn a discrete confidence threshold to identify the semantic predictions for each category. Both of these policies are trained by RL in low-dimensional discrete action space.
%we adopt low-dimensional discrete action space for each policy.
% \liudai{which can accommodate among different object categories, dynamically learns a discrete confidence threshold to verify the segmentation results for each object category. Compared with previous methods using preset hard threshold regardless of the agent's observation ~\cite{chaplot2020object, ramakrishnan2022poni, georgakis2022l2m}, our agent is more intelligent and wisely strict with a learning-based verification policy, which builds the second part of the pun in our title \textit{Think Twice}.}
Through experiments, the simultaneous two-policy mechanism and discrete action space design dramatically reduce the difficulty in learning for 3D-aware ObjectNav and achieve better performance than existing modular-based navigation strategies~\cite{ramakrishnan2022poni,luo2022stubborn}.

Through extensive evaluation on the public benchmarks, we demonstrate that our method performs online 3D-aware ObjectNav at 15 FPS while achieving the state-of-the-art performance on navigation efficiency. Moreover, our method outperforms all other modular-based methods in both efficiency and success rate with up to 30x times less computational cost.

Our main contributions include:

\begin{itemize}
    \item We present the first 3D-aware framework for ObjectNav task.
    
    \item We build an online point-based construction and fusion algorithm for efficient and comprehensive understanding of floor-level 3D scene representation.
    
    \item We propose a simultaneous two-policy mechanism which mitigates the problem of low sample efficiency in 3D-aware ObjectNav policy learning. 
    
\end{itemize}

\section{Related Work}
\label{sec:related}
\noindent\textbf{GoalNav with Visual Sequences.} There are constantly emerging researches on object goal navigation. One line of recent works directly leverages RGBD sequences, called end-to-end RL methods~\cite{wijmans2019dd}, which tends to implicitly encode the environment and predict low-level actions. These works benefit from visual representation~\cite{mousavian2019visual, yang2018visual}, auxiliary task~\cite{ye2021auxiliary}, and data augmentation~\cite{maksymets2021thda}, demonstrating strong results on object goal navigation benchmarks~\cite{batra2020objectnav, habitatchallenge2022}. 
However, aiming to learn all skills through one policy from scratch, e.g., avoiding collisions, exploration, and stopping, it's well known that end-to-end RL methods suffer from low sampling efficiency for training and limited generalizability when transferred to the real world~\cite{ramakrishnan2022poni,campari2020exploiting}.
Instead, our work uses explicit map to represent the environment, which ensures our sample efficiency and also obtain more generalizability through a modular-based paradigm~\cite{ramakrishnan2022poni, batra2020objectnav}.

%it's well known that it suffers from apparent performance drop when generalizing to novel or real-world scenes~\cite{chaplot2020object,campari2020exploiting,maksymets2021thda,ramakrishnan2022poni}.

\noindent\textbf{GoalNav with Explicit Scene Representations.} 
To ease the burden of learning directly from visual sequences, another category of methods, called modular-based methods~\cite{chaplot2020learning,chaplot2020object,Parisotto2018NeuralMS,Gupta2017CognitiveMA,Georgakis2019SimultaneousMA}, use explicit representations as a proxy for robot observations.
By leveraging explicit scene representations like scene graph~\cite{zhu2021soon, qiu2020learning} or 2D top-down map~\cite{ramakrishnan2022poni, georgakis2022l2m}, modular-based methods benefit from the modularity and shorter time horizons. They are considered to be more sample efficient and generalizable ~\cite{ramakrishnan2022poni,georgakis2022l2m}.
Recent progress in modular-based methods has proposed a frontier-based exploration strategy~\cite{ramakrishnan2022poni}, a hallucinate-driven semantic mapping method~\cite{georgakis2022l2m}, and novel verification stage~\cite{luo2022stubborn}. In contrast with prior map-based works, our method utilizes 3D spatial knowledge, including 3D point semantic prediction and consistency, enabling a more comprehensive understanding of the environment. 

\begin{figure*}[t]
\centering
\begin{overpic}
[width=\linewidth]
%[width=\linewidth,grid,tics=10]
{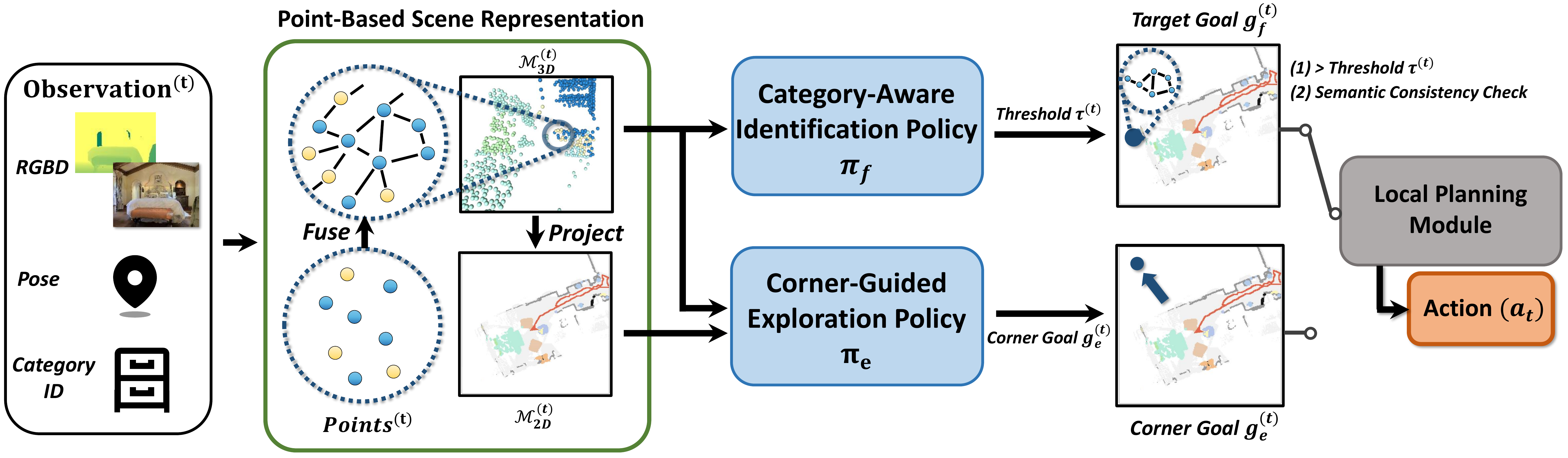}
    % \put(12,30.5){\footnotesize (a)}
    % \put(38,30.5){\footnotesize (b)}
    % \put(65,30.5){\footnotesize (c)}
    % \put(87.5,30.5){\footnotesize (d)}
\end{overpic}
\caption{
An overview of our framework. We take in a posed RGB-D image at time step $t$ and perform point-based construction algorithm to online fuse a 3D scene representation ($\M_{3D}^{(t)}$), along with a $\M_{2D}^{(t)}$ from semantics projection. Then, we simultaneously leverage two policies, including a \textit{corner-guided exploration policy} $\pi_e$ and \textit{category-awre identification policy} $\pi_f$, to predict a discrete corner goal $g_e^{(t)}$ and a target goal $g_f^{(t)}$ (if exist) respectively. Finally, the local planning module will drive the agent to the given target goal $g_f^{(t)}$ (top priority) or the corner goal $g_e^{(t)}$.
}
\vspace{-0.4cm}
\label{fig:overview}
\end{figure*}
% (a). \textit{3D Points Fusion.} Based on the back-projection points, we perform 3D points fusion to online organize the 3D points. (b). \textit{Exploration Policy and verification Policy} takes both 2D map (from semantic mapper) and 3D map to predict a discrete direction. \textit{V} takes 3D map and predict a confidence threshold with a label consistency mechanism to determine a final target goal. (c) Finally, a \textit{ local planner} outputs a moving action to reach the target goal/direction. 

\noindent\textbf{Embodied AI tasks with 3D Scene Representation.}
There are considerable research leveraging 3D scene representation on certain embodied AI tasks, e.g., object grasping~\cite{cao2021suctionnet, choi2018learning}, drawer opening~\cite{mu2021maniskill, shen2022learning}.
These works leverage various routes, including reinforcement learning~\cite{gadre2021act}, imitation learning~\cite{shen2022learning}, and supervised learning~\cite{cao2021suctionnet} with 3D scene representation, such as mesh, dense grids. However, most of these 3D-aware embodied AI tasks only perform in a limited space~\cite{mu2021maniskill, shen2022learning, choi2018learning}, \textit{e.g.}, near one table or drawer. Under large scale environments, such as floor-level scenes in ObjectNav, the existing methods would suffer from complex 3D observation and large computational costs. In this work, we propose a framework through leveraging a point-based construction module and two dedicatedly designed exploration and identification policies, to enable a 3D-aware agnet for ObjectNav.

\section{Method}
\label{sec:method}
\subsection{Task Definition and Method Overview}
% \qingnan{In an unknown environment, the Object Goal Navigation task requires the agent to navigate to an instance of the specified target category. For fair comparison, we follow the previous problem setting \cite{savva2019habitat,habitatchallenge2022}. As initialization, the agent is located randomly without the access to a pre-built environment map, and provided with a target category ID. At each time step, the agent receives the noiseless onboard sensor readings, including an egocentric RGB-D image and a 3-DoF pose (2D position and 1D rotation) relative to the beginning of the episode, then estimates its action for movement in a discrete action space, consisting of \texttt{move\_forward}, \texttt{turn\_left} and \texttt{turn\_right}. Given a limited time budget of 500 steps, the agent terminates the movement until it is within 1 meter of an object of the specified category.
% % Figure \ref{fig:overview} provides an overview of the proposed 3D-aware object-goal navigation algorithm. At each time step, provided the input observation from the onboard sensor, we first construct the 3D scene representation via the 3D point fusion module (Sec.\ref{sec:pointsfusion}).
% }

\noindent\textbf{Object Goal Navigation Task.} 
In an unknown environment, the Object Goal Navigation task requires the agent to navigate to an instance of the specified target category. For fair comparison, we follow the previous problem setting \cite{savva2019habitat,habitatchallenge2022}. As initialization, the agent is located randomly without access to a pre-built environment map, and provided with a target category ID. At each time step $t$, the agent receives noiseless onboard sensor readings, including an egocentric RGB-D image and a 3-DoF pose (2D position and 1D orientation) relative to the starting of the episode. Then the agent estimates its action $a_t \in \A$ for movement in a discrete action space, consisting of \texttt{move\_forward}, \texttt{turn\_left}, \texttt{turn\_right} and \texttt{stop}. Given a limited time budget of 500 steps, the agent terminates the movement until it is within 1 meter of an object of the specified category.

% \refine{We follow the definition and setting of the object-goal navigation task as described in ~\cite{savva2019habitat,habitatchallenge2022}}. The agent is expected to navigate to an instance of a specific object category (e.g., \textit{chair}) in an unknown environment. The agent is provided with a target category object ID and initialized at a random location without the access to a pre-built environment map. At each time step $t$, the agent first receives onboard sensor readings, including an egocentric RGB-D image from an RGBD camera along with 2D location and 1D orientation information (relative to the starting point) from GPS and compass;
% and then it executes an action $a_t \in \A$, where the action space $\A$ consists of four discrete actions: \texttt{move\_forward}, \texttt{turn\_left}, \texttt{turn\_right} and \texttt{stop}. 
% \sout{In 500 steps, the agent is required to navigate to the target object within $d_s=1.0\text{m}$  and then take the \texttt{stop} action to complete the task.}
% \liudai{Given a limited time budget(500 steps in our setting), the agent is expected to navigate and finally call \texttt{stop} within $d_s=1.0\text{m}$ from the target object.}

%and must navigate using its (noiseless) onboard sensors including RGBD cameras and GPS+compass (location and orientation relative to start of episode). And the action space is discrete and 

\noindent\textbf{Method Overview.} 
Figure~\ref{fig:overview} provides an overview of the proposed 3D-aware ObjectNav method. 
Our method takes RGBD frames along with pose sensor readings as input, to online construct a point-based scene representation $\M_{3D}$ (Sec.~\ref{sec:pointsfusion}), which is further projected to construct a 2D semantic map $\M_{2D}$. Given the structured 3D points $\M_{3D}$ and 2D map $\M_{2D}$, our framework simultaneously performs two complementary policies (Sec.~\ref{sec:rl}), the \emph{exploration} policy and \emph{identification} policy at a fixed time cycle of 25 steps. The exploration policy predicts a long-term discrete corner goal $g_e$, to drive the agent to explore the surrounding environment. Meanwhile, the identification policy evaluates the 3D points $\M_{3D}$ at each step and outputs a target object goal $g_f$ if its semantic prediction is confident and consistent.
The $g_f$ will be set as the approaching target for the agent once it exists, otherwise the agent will navigate to the long-term corner goal $g_e$ .
An underlying local planning module will navigate the agent towards the \emph{goal} using analytical path planning.
\subsection{ Navigation-Driven 3D Scene Construction}
\label{sec:pointsfusion}
During navigation, the 3D-aware agent will
% \liudai{To enable a 3D-aware ObjectNav framework, we expect our agent could} 
constantly obtain new observations and incrementally build a fine-grained 3D scene representation, integrating spatial and semantic information to drive the agent.
% Therefore, the online 
%  \refine{\sout{constructing 3D scene representation for GoalNav in active learning-based methods is fairly challenging due to two major requirements.}}
% However, given that our agent is deployed for a floor-level GoalNav task, building a qualified 3D scene representation is fairly challenging due to two major demands. First, it have to be efficient enough to online construct a 3D scene with minimal memory costs. Second,
% the scene representation should fuse the semantic predictions from multi-view observation of the agent and \warning{evaluate} the spatial semantic consistency of the target object.
However, given that our agent is deployed for a floor-level GoalNav task, it is fairly challenging to construct and leverage 3D representation across the entire scene while keeping an acceptable computational cost. 
Accordingly in this section, 
% \sout{First, it have to be efficient enough to online construct the 3D scene geometry and with minimal memory costs. Second, the scene representation should fuse the semantic predictions from multi-view observation of the agent and \warning{evaluate} the spatial semantic consistency of the target object.}
we extend an online point-based construction algorithm~\cite{zhang2020fusion} to online organize the 3D points and further empower semantic fusion and consistency estimation. This design is tailored for a comprehensive scene understanding of the ObjectNav agent, requiring little computational resources.
% \sout{tailored for navigation policy learning.}

% The algorithm ~\cite{zhang2020fusion} leverages the way introduced by a point-based fusion framework ~\cite{zhang2020fusion} to online organize the 3D points and further devise the semantic fusion and consistency estimation module tailored for navigation policy learning.

\noindent\textbf{3D Scene Representation.}
At time step $t$, we represent the 3D scene $\M_{3D}$ as the point clouds, denoted as $P^{(t)} = \{(P_l^{(t)}, P_s^{(t)}, P_c^{(t)})\} \in \R^{N^{(t)} \times (M+4)}$, where $N^{(t)}$ is the point number. For each point $i$, the $M+4$ channels include the point position $P^{(t)}_{i,l} \in \R^{3}$, point semantics $P^{(t)}_{i,s} \in \R^{M}$ and the point-wise spatial semantic consistency information $P^{(t)}_{i,c} \in \R^{1}$.

% At time step $t$, we are provided with the temporally accumulated sequence of RGB image, depth image and 3-DoF pose $\{I^{(k)}, D^{(k)}, C^{(k)}\}_{k=1}^{t}$. We obtain the point position $P_l^{(t)}$ by back-projecting all the depth images into the 3D world space via their corresponding poses. To obtain the point semantics $P_s^{(t)}$, we leverage the pretrained MaskRCNN \cite{he2017mask} or Rednet \cite{jiang2018rednet} (according to the evaluation benchmark) to obtain the pixel-wise semantic predictions from the RGB image, which are further propagated to the corresponding 3D point cloud.

\noindent\textbf{Online 3D Point Fusion}
%  \sout{Given a sequence of posed RGB images $I^c_t$ and depth images $I^d_t$ at time step $t$ during navigation,}
Given a new captured posed RGB image $I_c^{(t)}$ and depth image $I_d^{(t)}$ at time step $t$, the agent can obtain the point position $P_l^{(t)}$ by back-projecting all the depth images into the 3D world space via their corresponding poses.
These points will be organized by a point-based construction algorithm~\cite{zhang2020fusion}.Here, we briefly revisit this strategy.

\begin{figure}[t]
\centering
\begin{overpic}
[width=1.0\linewidth]
%[width=\linewidth,grid,tics=10]
{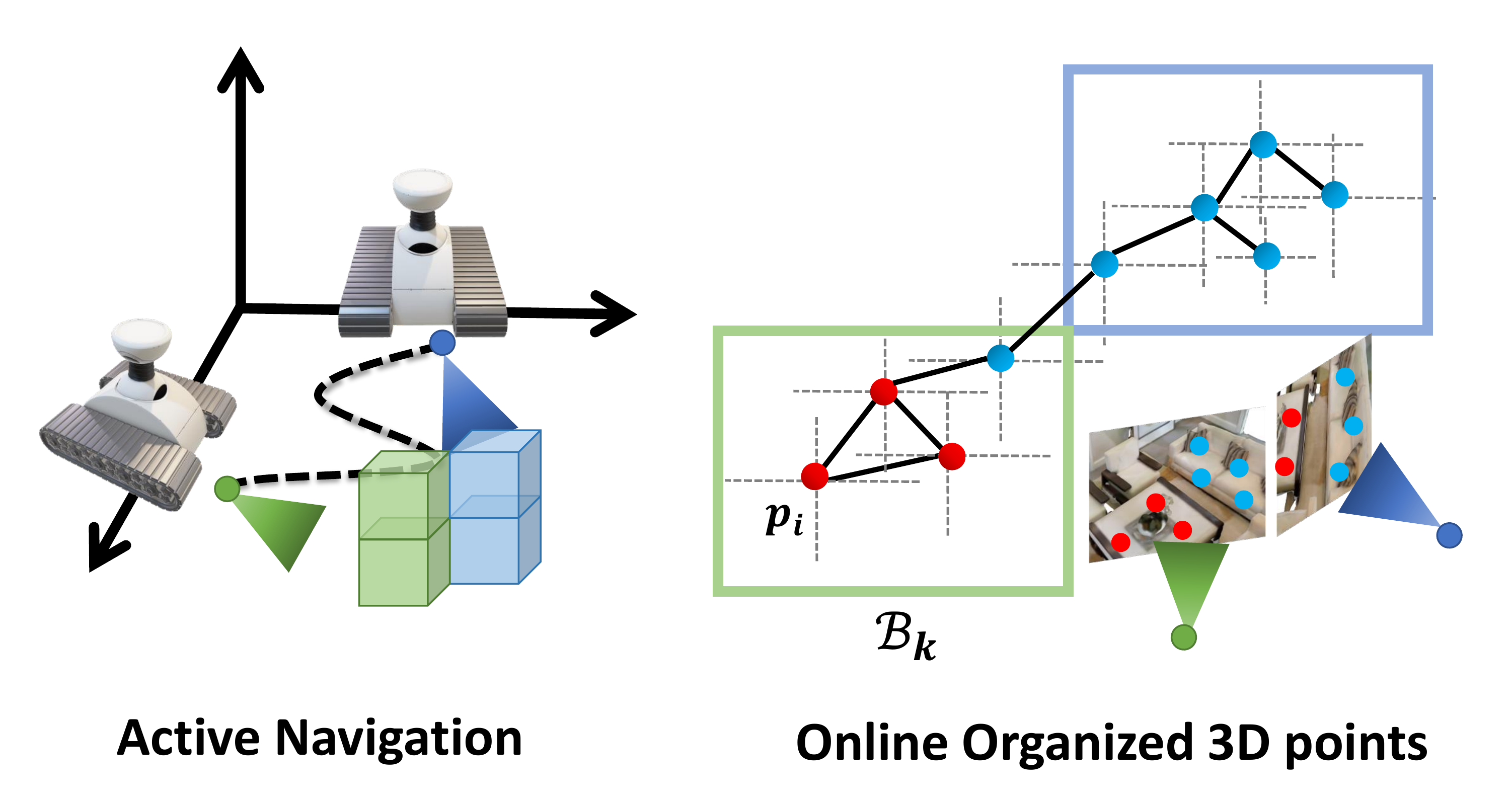}
    % \put(12,30.5){\footnotesize (a)}
    % \put(38,30.5){\footnotesize (b)}
    % \put(65,30.5){\footnotesize (c)}
    % \put(87.5,30.5){\footnotesize (d)}
\end{overpic}
\caption{
Illustration of online 3D point fusion. \textbf{(Left)} A robot takes multi-view observations during navigation. \textbf{(Right)} The points $p$ are organized by dynamically allocated blocks $\B$ and per-point octrees $\bO$, which can be used to query neighborhood points of any given point.
}
\vspace{-0.4cm}
\label{fig:map}
\end{figure}

The construction algorithm dynamically allocates occupied 3D blocks $\{\B_k\}$ along with their index $k$ maintained by a tree-based method~\cite{jagadish2005idistance}.   
Each block $\B_k$ is defined by the boundary of constant length ($10$cm) along the X, Y and Z axes, \textit{e.g.}, $[X_{min}(B_k), X_{max}(B_k)]$. And the points $p_{l,x}\in$ $[X_{min}(B_k), X_{max}(B_k)]$ (the same requirement holds for Y and Z axes) be recorded by the block $\B_k$.
% records the points inside it, \textit{i.e.} the points within a given coordinate range:
% \begin{equation}
% \label{equ:blocks}
%   x_p\in[X^{min}_k, X^{max}_k), y_p\in[Y^{min}_k, Y^{max}_k), z_p\in[Z^{min}_k, Z^{max}_k),
% \end{equation}
%multi pairs composed of a 3D block $\B_k$ and its corresponding indexing $k$. Each block 
Given any 3D point $p_i$, the algorithm can achieve efficient neighborhood retrieval with the corresponding block index $k$.
Furthermore, a one-level octree $\bO_i$ for each point $p_i$ is constructed to obtain the fine-grained spatial information among points. Specifically, we connect each point with its nearest points in the eight quadrants of the Cartesian coordinate system (See Figure~\ref{fig:map}).
Powered by this point-based construction strategy, give any point, we can efficiently querying this point with it's neighbor points by blocks retrieval and octree.
This algorithm for organizing 3D points can run at 15 FPS while requiring reasonable memory resources (about $\sim500$ MB for one entire scene).  We provide more detailed description in the the supplemental material.

% Note that, \refine{\sout{under} \liudai{for} ObjectNav task}, there are \liudai{often} considerable overlaps between consecutive frames, \sout{Therefore,} \liudai{so} we can reuse most of ($\sim60\%$) the blocks \sout{which} \liudai{and} significantly boost the running efficiency. Additionally, we only insert the newly observed points that have a distance (greater than 4 cm) from all existing \sout{points} \liudai{ones}, \sout{making} \liudai{urging} the points \liudai{to distribute} as \sout{uniformly distributed} \liudai{uniform} as possible. 

% \vspace{1em}
% \input{fig/consistency}
\noindent\textbf{Online Semantic Fusion.} 
% \refine{
% \sout{One of the main advantages of using 3D-aware scene representation is to fuse multi-view semantic predictions to} 
With an efficient reconstruction algorithm in hand, we can directly fuse temporal information, \textit{e.g.,} multi-view semantic predictions, to achieve more accurate and consistent scene understanding. Specifically, any point $p_i$ which has been captured by a sequence of RGBD frames $\{I_c^{(t)},I_d^{(t)}\}$ could have multiple semantic predictions $\{p_{i,s}^{(t)}(I_c^{(t)})\}$. We thus propose to online aggregate the multi-view 2D semantic predictions using a max-fusion mechanism to obtain the final 3D semantic prediction:
% \sout{Here, we use a pre-trained 2D model ~\cite{jiang2018rednet}, following existing works~\cite{ye2021auxiliary, ramakrishnan2022poni}}.
%Benefit from the multi-view observation, the semantic prediction of points can be on-the-fly updated during the navigation. 
% Considering a sequence of RGBD observations ($I^c_{t=1..N}$ and $ I^d_{t=1..N}$), our method first obtains the semantic prediction $S_{2D}(p_i|I^c_t)$ by a pre-trained 2D network ~\cite{jiang2018rednet}, following existing works~\cite{ye2021auxiliary, ramakrishnan2022poni}. 
% \he{check:} Note that this semantic prediction is from single view and thus highly error-prone, however previous works simply feed this prediction to the navigation policy, at the risk of falsely guiding the goal navigation.
% Leveraging our 3D sparse map, we can easily fuse the predictions to lower the errors. For any 3D point, Our 3D sparse map enables us to efficiently 1) find the corresponding pixels cross multi-view observation and 2) search its nearest neighbor points, which enables us to perform the semantic fusion.
%  $p_{i,s}(I_c_{(t)})$ from a pre-trained 2D segmentation model Mask R-CNN ~\cite{He2020MaskR} or RedNet~\cite{jiang2018rednet}, 
\begin{equation}
   p_{i,s}^{(t)} = \ \N(\max(\{p_{i,s}^{(t)}(I_c^{(t)}))\})),
\end{equation}
where the $\max$ is performed on each semantic category, followed by a normalization $\N$ to linearly scale the probability distribution.
Note that, the alternatives to fuse semantic predictions do exist, e.g. 3D convolution~\cite{huang2021supervoxel,liu2022ins}, Bayesian updating~\cite{mccormac2017semanticfusion}.
However, directly conducting 3D convolution into such a floor-level 3D representation would inevitably lead to a huge rise of computational cost, especially in the context of learning-based policy.
We find that maximizing the 2D semantic prediction can already achieve impressive improvement on semantic accuracy (see Figure~\ref{fig:result_consistency}), with higher memory efficiency and time efficiency.
Similar findings have also been reported and exploited in relevant works~\cite{chaplot2021seal, grinvald2019volumetric}.
% Moreover, through experiment, we find that the max-fusion demonstrates better performance than Bayesian-fusion~\cite{mccormac2017semanticfusion}. 

\noindent\textbf{Spatial Semantic Consistency.} Based on the fact that semantic label should remain consistent for all the points in a single object, we propose to calculate the spatial semantic consistency information $P^{(t)}_c$ as part of the navigation-driven 3D scene representation. To be specific, $P^{(t)}_{i,c}$ is computed as the maximum semantic KL-divergence between point $P^{(t)}_i$ and its octree $\bO(P^{(t)}_i)$:
\begin{equation} \label{equation:KL}
    P^{(t)}_{i,c} = max(\{KL(P^{(t)}_{i,s},P^{(t)}_{j,s}) | \forall P^{(t)}_j \in \bO(P^{(t)}_i) \}),
\end{equation}
where $KL$ denotes the KL-divergence computation, which is a statistical distance that measures the semantic probability distribution between $P^{(t)}_{i,s}$ and $P^{(t)}_{j,s}$. Note for point $P^{(t)}_i$, if we count all its spatially close points as the neighbourhood $\N(P^{(t)}_i)$, it could be time consuming to calculate Equation \ref{equation:KL}, and the spatially close points do not help relieve the issue of outlier points as mentioned above. Therefore, we use the pre-built octree $\bO_i$ to retrieval 8 nearest point in the quadrants of the Cartesian coordinate system.

\subsection{Simultaneous Exploration and Identification}
\label{sec:rl}
With the aggregated 3D information, we expect to empower a 3D-aware agent for the ObjectNav task. 
However, despite the efficient 3D scene representation, the agent still suffers from the complex and high-dimensional observations, leading to a lower sample efficiency in RL and hampering the navigation policy learning.
Therefore, we leverage two complementary sub-policies: corner-guided exploration policy and category-aware identification policy. Each policy learns to predict low-dimensional discrete actions and outputs a goal location to navigate the agent, resulting in a strong performance while requiring less training time. We will detail the two policies below.
% Furthermore, we design a policy with discrete action space for each sub-task, which significantly improves the sampling efficiency. 

\noindent\textbf{Observation Space.} At each time step $t$, both policies take fine-grained 3D observation $x_{3D}^{(t)} = \{P^{(t)} \in ((4+m) \times N)\}$ based on 3D scene representation $\M_{3D}$. Here, the $N$ indicates the point number (we sample 4096 points) and the $m+4$ channels are comprised of point position $p_l^{(t)} \in \R^3$, fused semantic predictions $p_s^{(t)}\in \R^m$ and spatial semantic consistency $p_c^{(t)} \in \R^1$.
Following existing works~\cite{chaplot2020learning,chaplot2020object}, we use an additional egocentric 2D map $\M_{2D}$ for exploration policy and the local path planning module, which is directly obtained by a project-to-ground operation. More detailedly, for 2D observation $x_{2D}^{(t)} \in ((2+m) \times M \times M)$ from 2D map $\M_{2D}$, the first two channels represent obstacles and explored area, and the rest of the channels each corresponds to an object category.
Here, $\M_{2D}$ (in a resolution of $M=240$ with $20 \text{cm}$ grids) is constructed to give a large perception view of the scene, while 3D points perform as a fine-grained observation of objects. In addition to the scene representations, we also pass the goal object category index $o_{ID}$ as the side input to both policies.
% The experiment shows that the combination demonstrates better performance than any individual representation (See Table. \ref{tab:ablation}).}
\begin{figure}
% \centering
% \vspace{-0.3cm}

\begin{overpic}
% []
%[width=\linewidth,grid,tics=10]
[width=1.0\linewidth]
{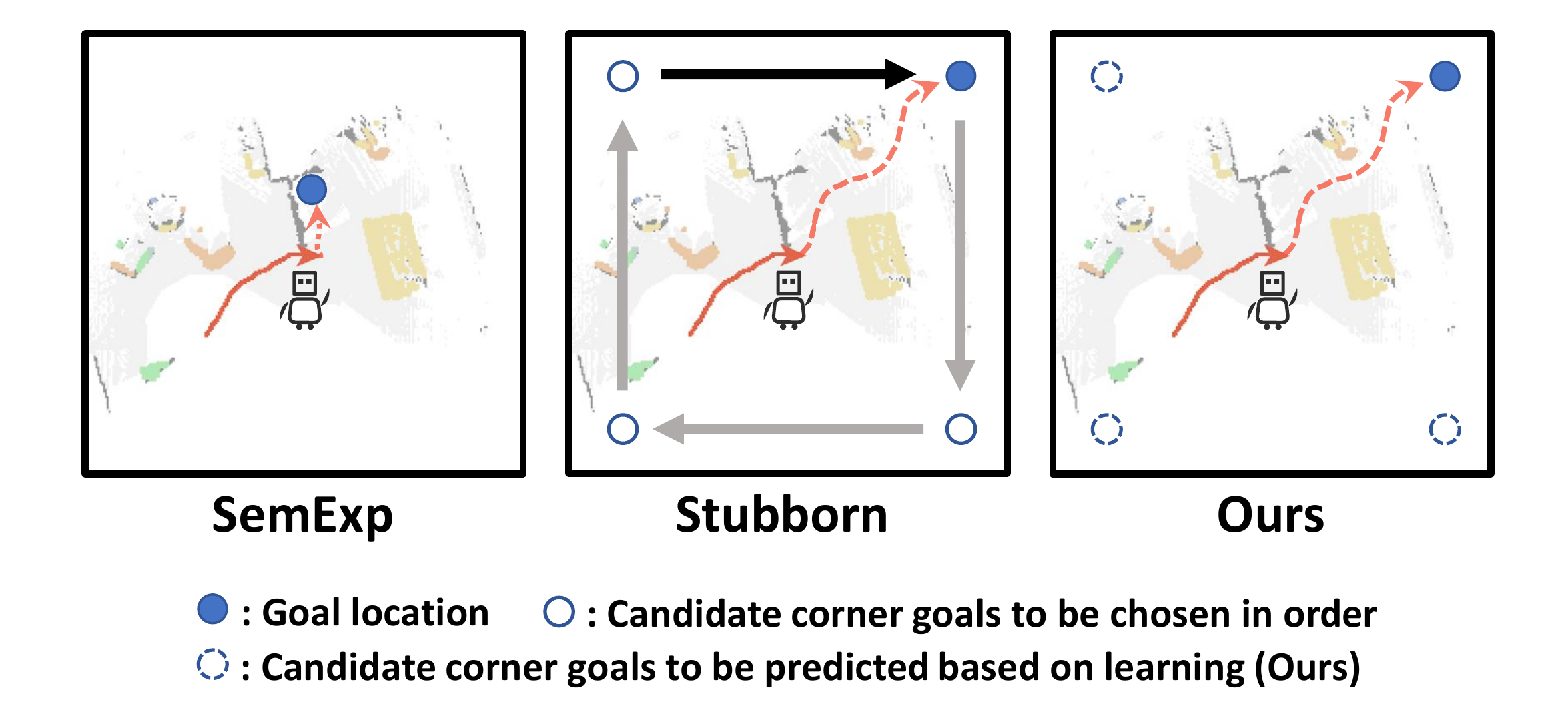}
    % \put(12,30.5){\footnotesize (a)}
    % \put(38,30.5){\footnotesize (b)}
    % \put(65,30.5){\footnotesize (c)}
    % \put(87.5,30.5){\footnotesize (d)}
\end{overpic}
\caption{
Illustration of exploration policy. \textbf{(Left)} Learning-based continuous global goal~\cite{chaplot2020object}; \textbf{(Middle)} Heuristic direction selection~\cite{luo2022stubborn}; \textbf{(Right, ours)} Learning-based corner goal prediction.
}
\vspace{-0.6cm}
\label{fig:corner_goal}
\end{figure}

\noindent\textbf{Corner-Guided Exploration Policy.}
The exploration policy attempts to guide the agent to explore and perceive the surrounding environment where it could access any instance of the target object category.
We observe that existing learning-based exploration policies predict goal locations over the 2D map in continuous or large-dimensional discrete action space (Figure~\ref{fig:corner_goal} Left), suffering from low sample efficiency.
Therefore, we define a corner-guided exploration policy $g_e = \pi_{e}(x_{3D}, x_{2D}, o_{ID};\theta_e)$ that predicts a corner goal $g_e$ to drive the agent(Figure~\ref{fig:corner_goal} Right). Here, the $\theta_e$ indicates the parameters of the policy,
and $g_e$ is one of the four pre-defined corner goals \{\texttt{Top Left}, \texttt{Top Right}, \texttt{Bottom Left}, \texttt{Bottom Right}\} of the 2D map. 

Compared to predicting goals in a continuous or high-dimensional action space, learning to predict the four corner goals significantly reduces the learning difficulty. Moreover, as noted by previous studies~\cite{luo2022stubborn, cao2021tare}, the corner-goal-based exploration strategy exhibits the capacity to achieve efficient exploration through avoiding back-and-forth pacing.
Superior to using other heuristic corner goal exploration strategies (Figure~\ref{fig:corner_goal} Middle), our agent can learn from the 3D scene priors to behave more intelligently. Demonstrations of our corner-guided exploration can be found in the attached video.

\begin{figure}
% \centering
% \vspace{-0.3cm}

\begin{overpic}
% []
%[width=\linewidth,grid,tics=10]
[width=1.0\linewidth]
{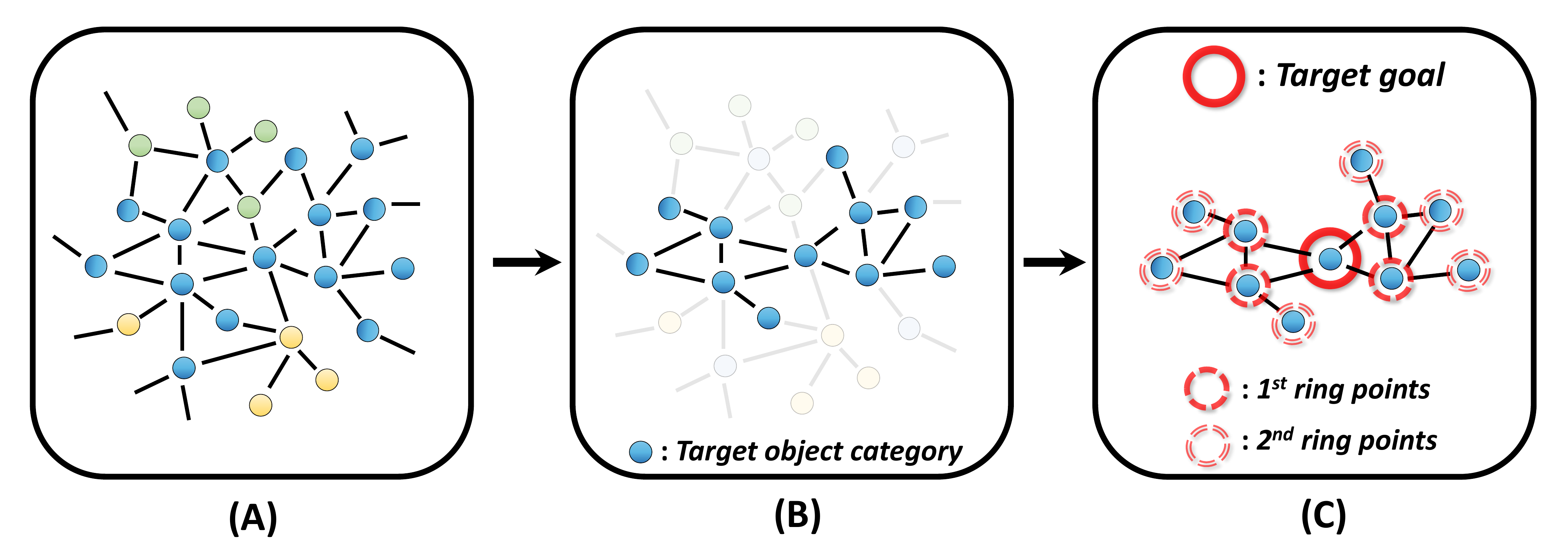}
    % \put(12,30.5){\footnotesize (a)}
    % \put(38,30.5){\footnotesize (b)}
    % \put(65,30.5){\footnotesize (c)}
    % \put(87.5,30.5){\footnotesize (d)}
\end{overpic}
\caption{Illustration of identification policy. From A $\rightarrow$ B, fused points are filtered by the category-aware predicted threshold $\tau$. 
From B $\rightarrow$ C, the policy further checks the spatial label consistency of the points and identifies the target goal.
}
\vspace{-0.4cm}
\label{fig:identification}
\end{figure}

\label{method:ver}
\noindent\textbf{Category-Aware Identification Policy.}
During navigation, the agent consistently makes semantic predictions to identify an instance of target object category. 
% Most works~\cite{chaplot2020object, georgakis2022l2m} simply use a preset hard confidence threshold on semantic predictions to \warning{identify} the target object goal.
% However, this strategy has two major limitations: 1) The semantic prediction is error-prone across different categories and different camera views;
% 2) It ignores the consistency of the semantic prediction in 3D space
Most works ~\cite{chaplot2020object, georgakis2022l2m} simply use a preset hard confidence threshold for target identification.
However, this strategy is inherently sub-optimal due to the considerable variability in semantic prediction results across different categories and observation angles. As a result, a preset threshold would be unable to adequately adapt to the ever-changing nature of these scenarios.
Also, it ignores to consider the consistency of the semantic prediction in 3D space.
% These limitations can lead to numerous mistakes, and what's worse, the agent won't be able to recover from the mistake once it starts marching to the wrong goal.

To tackle this issues, we propose to leverage both dynamic confidence threshold and spatial semantic label consistency for target identification.
We define a policy $s = \pi_{f}(x_{3D},o_{ID}; \theta_f)$ which takes the 3D observation $x_{3D}$ and target category index $o_{ID}$ and outputs a threshold-indicating action $s \in \{0,1...,9\}$. And the dynamic threshold $\tau$ can be obtained by:
\begin{equation}
    \tau = \tau_{low} + s \cdot \frac{1-\tau_{low}}{10},
\end{equation}
where the $\tau_{low}$ is set to $0.5$ in our implementation for a threshold range $\tau \in [0.5, 0.95]$. 
The $\tau$ will be used to dynamically identify the points belonging to the target object (Figure~\ref{fig:identification} Middle).
% The dynamic confidence threshold $\tau$ is conditioned on the target category index $o_{ID}$ and so far fused 3D points $\M_{3D}$, filter out the . 
It is worth mentioning that that this policy also utilizes a low-dimensional discrete action space, which is fairly easy for the agent to learn.

To obtain the final target goal $g_f$, our method further checks the spatial semantic label consistency. Specifically, we use the points $\{p_i|(p_i,p)\in\bO_p\}$ connected by the per-point octree $\bO_p$ to approximately represent the 3D surface of the target object. Our insight is that the points along the target's surface should have consistent semantic labels. Therefore, we only identify those points who have at least 2-ring neighbors across the octrees $\{p_i|(p_i,p_j)\in\bO_{p_j}|(p_j,p)\in\bO_{p}\}$ as the target object goal $g_f$ (Figure~\ref{fig:identification} Right). See Figure~\ref{fig:identification} for visualized illustration and more details can be found in supplemental material.

\noindent\textbf{Local Planning Module.} 
The goals $g_e$ and $g_f$ from two polices will be consistently updated during navigation. 
Our method will preferentially utilize the target goal $g_f$ if it exists, otherwise take the long-term corner goal $g_e$ to explore.
% Here to minimize the steps required to find the target object, our method preferentially utilizes the target goal $g_f$ (if it exists), otherwise uses the long-term corner goal $g_e$ to explore.
To navigate to the given location, we use the Fast Marching Method~\cite{Sethian1999FastMM} to analytically plan the shortest path from the agent location. The agent then takes deterministic actions to follow this path.
% \warning{There are also some other learning-based alternatives for path planning, such as \cite{chaplot2020learning, wijmans2019dd}.}

\noindent\textbf{Rewards.} For the exploration policy, we share a similar reward design as~\cite{ye2021auxiliary,batra2020objectnav}. The agent receives a sparse success reward $r_{success} = 2.5$, a slack reward $r_{slack}=10^{-2}$ and an exploration reward $r_{explore}$. The exploration reward is a dense reward, defined by the number of new inserted point $n^{new}_p$ as $r_{explore} = n^{new}_p \times 10^{-3}$. The slack reward and exploration reward encourage the agent to take the most effective direction to the unobserved area. And for the identification policy, we combine the same success reward and slack reward borrowed from the exploration policy.

%But we believe that the confidential of agent to confirm the object should vary from the object category, because the semantic predictor is more accurate under common category  \textit{e.g.} chair, sofa. So we define a policy $\pi_{et}$ which output the threshold from observing the online capture points. 

\section{Experiments}
\label{sec:result}
\subsection{Experiment Setup.}
\begin{figure*}[t]
\centering
\begin{overpic}
[width=0.9\linewidth]
%[width=\linewidth,grid,tics=10]
{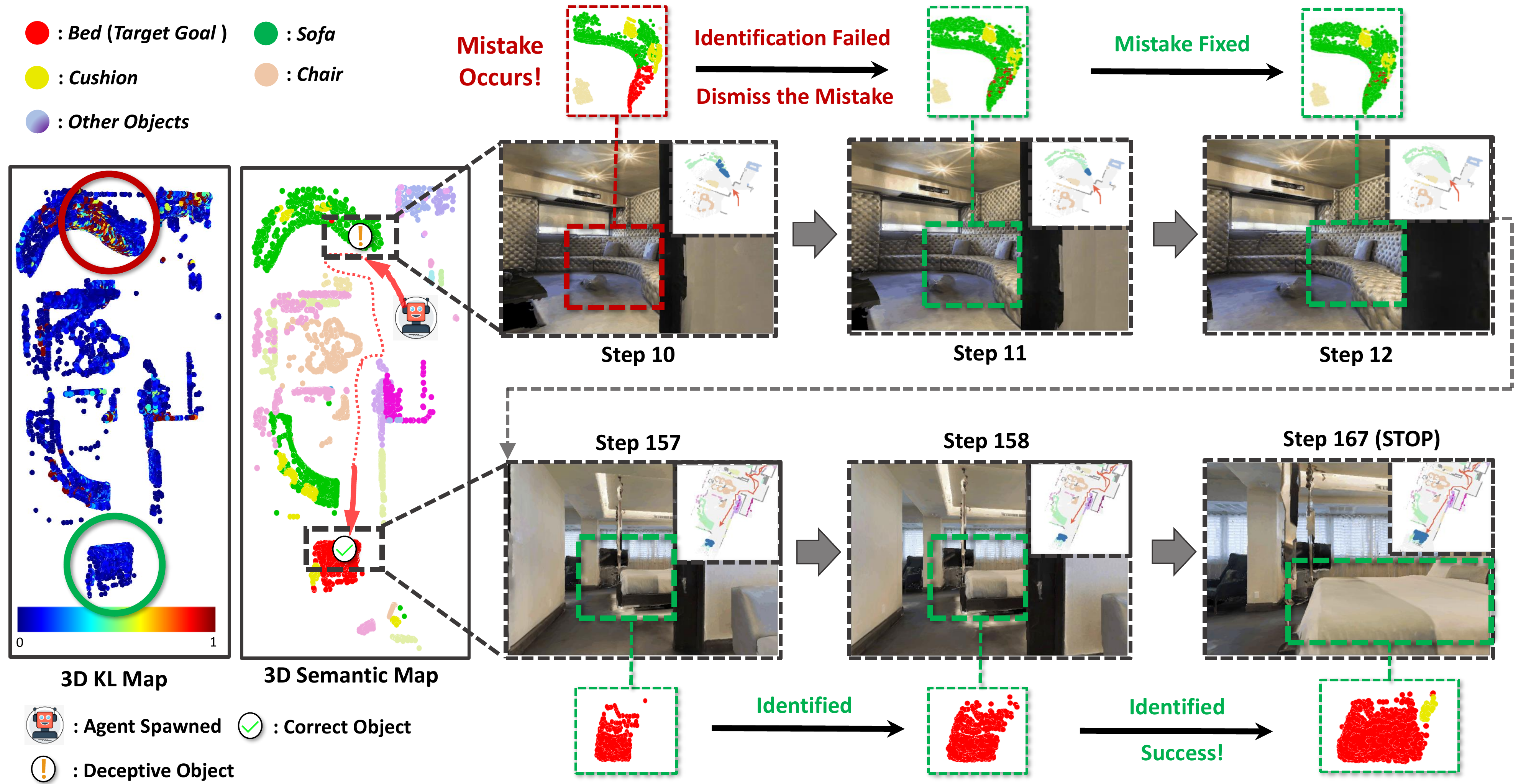}
    % \put(12,30.5){\footnotesize (a)}
    % \put(38,30.5){\footnotesize (b)}
    % \put(65,30.5){\footnotesize (c)}
    % \put(87.5,30.5){\footnotesize (d)}
\end{overpic}
\caption{
An qualitative visualization of the trajectory of the proposed method. We visualize an episode from MP3D where an agent is expected to find a \textit{bed}. The semantic prediction $p_s$ and spatial semantic consistency $p_c$ of points are visualized on the left. During navigation, the agent can successfully dismiss the wrong prediction and approach and finally call stop around the target object.
}

% An qualitative result of ObjectNav using the proposed method. We visualize an episode on Matterport3D (val), along with a top-down semantic map and KL map. The robot is expected to find a bed.
% At step 10, the robot obtain a partial observation of a sofa, which is mistakenly recognized as a bed. And during the approach to the sofa, the multi-view observation improve the semantic prediction and successfully dismiss the mistake. Then the robot continue to explore the environment until step=157, when the robot obtain new partial observation of an object which could be the bed. As the robots moving to the target, the points are more complete and the semantic prediction is tend to be bed. And for KL map, we can find that the false target goal (red circle) has large inconsistency contrasted to the correct object(green circle). 
%Please see supplementary or appendix for more examples.
% Here, the agent successfully dismissed a deceptive object \textbf{\textit{sofa}}  and finally stop at correct category object \textbf{\textit{bed}}.
%
% The visualization of one specific episode using the 3D key-points we maintain, during which the agent successfully dismissed a deceptive object \textbf{\textit{sofa}}  and finally arrived at its true goal object \textbf{\textit{bed}}. Besides the top-down semantic map, we also visualize the KL map, displaying the lay-out of inconsistency across the scene. The region where the deceptive \textbf{\textit{sofa}} is settled reaches the peak of KL-Divergence, which means this region is most likely to be avoided by our agent in verification stage.
\label{fig:result}
\end{figure*}
We perform experiments on the Matterport3D (MP3D)~\cite{chang2017matterport3d} and Gibson~\cite{xia2018gibson} datasets with the Habitat simulator~\cite{habitat19iccv}. Both Gibson and MP3D contain photorealistic 3D reconstructions of real-world environments.
For Gibson, we use 25 train / 5 val scenes from the Gibson tiny split. And we follow the same setting as in \cite{chaplot2020object,ramakrishnan2022poni} where we consider 6 goal categories, including \textit{chair}, \textit{couch}, \textit{potted plant}, \textit{bed}, \textit{toilet} and \textit{TV}. 
For MP3D, we use the standard split of 61 train / 11 val scenes with Habitat ObjectNav dataset ~\cite{savva2019habitat}, which consists of 21 goal categories (the full list can be found in the supplemental material).
%We follow the definition of the object-goal navigation task as described in ~\cite{batra2020objectnav}. 
Note that, the RGB-D and pose readings are noise-free from simulation (follow the definition of \cite{batra2020objectnav}). Estimation of the pose from noisy sensor readings is out of the scope of this work and can be addressed if necessary, by incorporating off-the-shelf robust odometry~\cite{Zhao2021TheSE, zhang2022asro}.

% For Gibson experiments, For MP3D experiments, we use the Habitat ObjectNav dataset ~\cite{savva2019habitat}, which consists of 21 goal categories (the full list can be found in the appendix).

% the effect of the 2D backbone

\noindent\textbf{Implementation Details.} On MP3D, we use the same pre-trained 2D semantic model RedNet~\cite{jiang2018rednet} as~\cite{ramakrishnan2022poni,ye2021auxiliary}. On Gibson, we leverage a Mask R-CNN~\cite{He2020MaskR}, which is trained with COCO dataset~\cite{Lin2014MicrosoftCC}. For each frame, we randomly sample 512 points for point-based construction. Moreover, we use PointNet~\cite{Qi2017PointNetDL} and fully convolutional networks~\cite{long2015fully} to obtain the feature of 3D points and the 2D map, respectively. During training, we sample actions every 25 steps and use the Proximal Policy Optimization (PPO) ~\cite{Schulman2017ProximalPO} for both exploration and identification policies. More implementation details can be found in the supplemental material.

\noindent\textbf{Evaluation Metrics.} Following existing works~\cite{Batra2020ObjectNavRO,ramakrishnan2022poni,georgakis2022l2m}, we adopt the following evaluation metrics: 
1) SPL: success weighted by path length. 
It measures the efficiency of the agent over oracle path length, 
which serves as the primary evaluation metric for Habitat Challenge~\cite{habitatchallenge2022}. 
2) Success rate: the percentage of successful episodes 
3) Soft SPL: a softer version of SPL measure the progress towards the goal (even with 0 success). 
4) DTS: geodesic distance (in m) to the success at the end of the episode.

% \subsection{Baselines}
\noindent\textbf{Baselines}. We consider mainstream baselines in the ObjectNav task. For end-to-end RL methods, we cover DD-PPO~\cite{wijmans2019dd}, Red-Rabiit~\cite{ye2021auxiliary}, THDA~\cite{maksymets2021thda}, and Habiat-Web~\cite{ramrakhya2022habitat}. For modular based methods, we cover FBE~\cite{Robotics1997Proceedings1I}, ANS~\cite{chaplot2020learning}, L2M~\cite{georgakis2022l2m}, SemExp~\cite{chaplot2020object}, Stubborn~\cite{luo2022stubborn} and PONI~\cite{ramakrishnan2022poni}. Note that, some works use additional data to improve the performance, \textit{e.g.} Habitat-web leverages human demonstration trajectories, and THDA utilizes data augmentation. It is challenging to compare all the methods fairly. Therefore, we are particularly interested in the three most relevant baselines: SemExp, Stubborn, and PONI. These three methods share the same 2D semantic predictors~\cite{jiang2018rednet, He2020MaskR} as our method.

% Therefore, we are espcially interested in three most related baselines:

% \textbf{- Stubborn:} is a strong baseline which consists of a semantic-agnostic goal selection module (introduced in Sec.) and a heuristic multi-frame verification module. %By improving the exploration effiectly 

% \textbf{- SemExp:} is proposed by \cite{chaplot2020object}, which uses 2D map as input and directly outputs the global goal. This method is originally incorporated with Mask R-CNN. To be fair, we use a same pre-trained 2D semantic predictor RedNet~\cite{jiang2018rednet} as existing works \cite{ramakrishnan2022poni, luo2022stubborn}.

% \textbf{- PONI:} is a modular approach which obtain the goal by a potential function network. Specifically, PONI use a dataset of semantic maps form 3D semantic annotations to train the potential function network. Unlike existing RL-based policy method, PONI is a interaction-free method without the need of simulator for training. 

\subsection{Results}

\begin{table}[!t]\centering
\caption{
ObjectNav validation results on Gibson and MP3D. Our method is trained with 5 seeds and report the averaged performance. The best of all methods and the best of all modular-based methods are highlighted in \textbf{bold} and \underline{underline} colors, respectively. Note that Habitat-Web takes use of extra data.
}
\scalebox{0.73}{
\setlength{\tabcolsep}{0.3mm}{
\begin{tabular}{l|ccc|ccc}
\hline
                                        & \multicolumn{3}{c|}{Gibson (val)}                      & \multicolumn{3}{c}{Matterport3D (val)}               \\ \cline{2-7} 
Method                                  & SPL$(\%)\uparrow$ & Succ.($\%$)$\uparrow$ & DTS(m)$\downarrow$ & SPL$(\%)\uparrow$ & Succ.($\%$)$\uparrow$ & DTS(m)$\downarrow$ \\ \hline \hline
DD-PPO~\cite{wijmans2019dd}             & $10.7$            & $15.0$              & $3.24$             & $1.8$             & $8.0$               & $6.90$             \\
Red-Rabbit~\cite{ye2021auxiliary}       & $-$               & $-$                 & $-$                & $7.9$             & $34.6$              & $-$                \\
THAD~\cite{maksymets2021thda}           & $-$               & $-$                 & $-$                & $11.1$            & $28.4$              & $5.58$             \\
Habitat-Web~\cite{ramrakhya2022habitat} & $-$               & $-$                & $-$                & $10.2$            & $\mathbf{35.4}$     & $-$                \\ \hline
FBE~\cite{Robotics1997Proceedings1I}    & $28.3$            & $64.3$              & $1.78$             & $7.2$             & $22.7$              & $6.70$\\
ANS~\cite{chaplot2020learning}          & $34.9$            & $67.1$              & $1.66$             & $9.2$             & $27.3$              & $5.80$             \\
L2M*~\cite{georgakis2022l2m}             & $-$               & $-$                 & $-$                & $11.0$            & $32.1$              & $5.12$             \\
SemExp*~\cite{chaplot2020object}        & $39.6$            & $71.7$              & $1.39$             & $10.9$            & $28.3$              & $6.06$             \\
Stubborn*~\cite{luo2022stubborn}        & $-$               & $-$                 & $-$                & $13.5$            & $31.2$              & $5.01$             \\
PONI~\cite{ramakrishnan2022poni}        & $41.0$            & $73.6$              & $1.25$             & $12.1$            & $31.8$              & $5.10$             \\ 
Ours                                    & \underline{$\mathbf{42.1}$}   & \underline{$\mathbf{74.5}$}     & \underline{$\mathbf{1.16}$}    & \underline{$\mathbf{14.6}$}   & \underline{$34.0$}              & \underline{$\mathbf{4.74}$}    \\ \hline
\end{tabular}}}
\label{tab:gibson_mp3d}
\vspace{-3.5mm}
\end{table} 
\begin{table}[!t]\centering
\caption{
ObjectNav validation results on MP3D-L2M~\cite{georgakis2022l2m}.
}
% \vspace{-5pt}
\scalebox{0.95}{
\setlength{\tabcolsep}{0.6mm}{
\begin{tabular}{l|cccc}
\hline
\multicolumn{1}{c|}{}           & \multicolumn{4}{c}{MP3D-L2M}                                                      \\ \hline
Method                          & SPL$(\%)\uparrow$ & \multicolumn{1}{l}{SoftSPL$(\%)\uparrow$} & Succ.$(\%)\uparrow$ & DTS(m)$\downarrow$ \\ \hline \hline
SemExp~\cite{chaplot2020object} & $16.5$            & $-$                                       & $28.1$              & $4.848$            \\ 
L2M~\cite{georgakis2022l2m}     & $14.8$            & $20.0$                                    & $34.8$              & $3.669$            \\
Ours                            & $\mathbf{21.2}$            & $\mathbf{30.5}$         & $\mathbf{40.2}$              & $\mathbf{3.278}$   \\ \hline
\end{tabular}
}}
\vspace{-0.45cm}
\label{tab:l2m}
\end{table} 
\noindent\textbf{Comparison on MP3D and Gibson.} We evaluate our approach on MP3D (val) and Gibson (val) with other baselines, including end-to-end RL(rows 1 - 4) and modular-based methods(rows 5 - 10). Note that, SemExp and Stubborn did not report the results on MP3D validation, while L2M uses a self-made dataset MP3D-L2M based on MP3D and tests fewer categories than what we do. We therefore faithfully provide the results, denoted with *, by evaluating with their public available code.
% Here, the SemExp~\cite{batra2020objectnav} and Stubborn~\cite{luo2022stubborn} do not provide their
The results are demonstrated in Table~\ref{tab:gibson_mp3d}. On both datasets, our method achieves the state-of-the-art ObjcetNav efficiency (SPL) among all methods (2.6\% higher on Gibson dataset and $8.1$\% higher on MP3D). For the success rate, our method achieves the best results among all modular-based methods, showing comparable performance with additional annotation methods THAD~\cite{maksymets2021thda} and Habitat-web~\cite{ramrakhya2022habitat}. 
Especially, compared with the modular-based methods, SemExp, Stubborn, and PONI, which share the same 2D semantic predictor~\cite{jiang2018rednet} as ours, the results fairly demonstrate the superiority of our framework on both efficiency and success rate. 
We also provide the results validated on MP3D-L2M in Table~\ref{tab:l2m}.
% To directly compare with the SemExp~\cite{chaplot2020object} and L2M~\cite{georgakis2022l2m}, we evaluate our method on the MP3D (L2M) dataset used by L2M. The results can be found in Table.\ref{tab:l2m}. Here we achieve the best performance among all used metrics. 
\begin{table}
\vspace{-10pt}
\caption{Comparison of different exploration policies. Here, all methods share the same identification strategy from~\cite{chaplot2020object} for fair comparison.}
\scalebox{1.0}{
\setlength{\tabcolsep}{0.5mm}{
\begin{tabular}{l|ccc}
\hline
Method & SPL(\%) &  Succ.(\%) & DTS(m) \\
\hline \hline
    Learn Continuous Goal. 
        &  11.1    &  28.6  &  6.354  \\
    Learn dense Grid Goal.
        &  12.7    &  29.5  &  5.635  \\
    Learn 8 corner goal.
        &  12.9    &  30.7  &  5.112  \\
    Heuristic. 4 corner goal.
        &  13.5    &  33.0  &  4.995  \\
\hline
    Learn 4 corner goal. (Ours)
        &  \textbf{13.9}    &  \textbf{33.5}  &  \textbf{4.931}  \\ 
\hline
\end{tabular}
}}
\vspace{-0.5cm}
\label{tab:result_exploration}
\end{table} 

We also provide a qualitative visualization of MP3D episodes in Figure~\ref{fig:result}. Here, our method online updates the semantic prediction and successfully dismisses the wrong target goal. For more qualitative results, please refer to the supplemental material.

% Considering three particular methods: SemExp, Stubborn, and PONI, which share the same 2D semantic predictor as ours, we outperform these three on all metrics, clearly performing the superiority of our method among modular-based methods.

% Moreover, compared to the end-to-end RL-based methods, like Habitat-web~\cite{habitatchallenge2022} trained with extra human demonstration, our method still achieves more efficient navigation with $10\%$ higher SPL and competitive success rate. Still, the performance of our method on the success rate could be further improved with a more accurate 2D semantic predictor~\cite{Liu2021SwinTV} and training data~\cite{Zhou2018SemanticUO}. A qualitative visualization can be found Figure.\ref{fig:result}. Here, our method online updates the semantic prediction and successfully dismisses the wrong target goal. For more episode qualitative results, please refer to the appendix. 

% On the Gibson dataset, our method achieves comparable performance to other baselines. However, due to the different 2D semantic predictors for methods, it is unfair to compare the final performance. Here, we provide the results only as a reference.

% \input{table/study_stop.tex}

\noindent\textbf{Comparison on Exploration Policy.}
We conduct an experiment to verify the efficiency of our corner-guided exploration policy on MP3D. To remove the effect of the 2D semantic predictor and identification policy, all competitors share the same semantic predictor and a heuristic identification policy proposed in SemExp~\cite{chaplot2020object}. The results are reported in Table~\ref{tab:result_exploration}. Our corner-guided exploration policy outperforms the mainstream existing methods, including learning-based ones~\cite{chaplot2020learning,georgakis2022l2m} and heuristic ones~\cite{luo2022stubborn}.  Our findings indicate that the best performance is achieved through learning to predict discrete corner goals from the four corners of the scene. This suggests that the four-corner design, which benefits from a small, discrete action space, is already capable of efficiently guiding the agent in exploring the environment.

%We believe the 4 direction is a good trade off between the moving freedom and dimensional of action space.

\begin{table}
\vspace{-10pt}
\caption{Comparison on different identification policies.}
\scalebox{1.0}{
\setlength{\tabcolsep}{0.5mm}{
\begin{tabular}{l|cc|ccc}
\hline
\multirow{2}{*}{Method}        & \multicolumn{2}{c|}{Type} & \multicolumn{1}{l}{\multirow{2}{*}{SPL(\%)}} & \multicolumn{1}{l}{\multirow{2}{*}{Succ.(\%)}} & \multicolumn{1}{l}{\multirow{2}{*}{DTS(m)}} \\
& Repr.  & Thre. \\ 
\hline \hline
\multirow{2}{*}{Deterministic} 
    % &  2D  &  0.75  &  13.0  &  29.7  &  5.168  \\
    &  2D  &  0.85  &  12.8  &  30.1  &  5.151  \\
    % &  3D  &  0.75  &  13.7  &  32.3  &  4.179  \\ 
    &  3D  &  0.85  &  13.8  &  32.5  &  4.987  \\ 
\hline
Learning (Ours)
    & 3D   &  -     &  \textbf{14.6}  &  \textbf{34.0}  &  \textbf{4.749}  \\ 
\hline
\end{tabular}}}
\vspace{-0.2cm}
\label{tab:verification}
\end{table} 
\begin{figure}[t]
\centering
\begin{overpic}
[width=\linewidth]
%[width=\linewidth,grid,tics=10]
{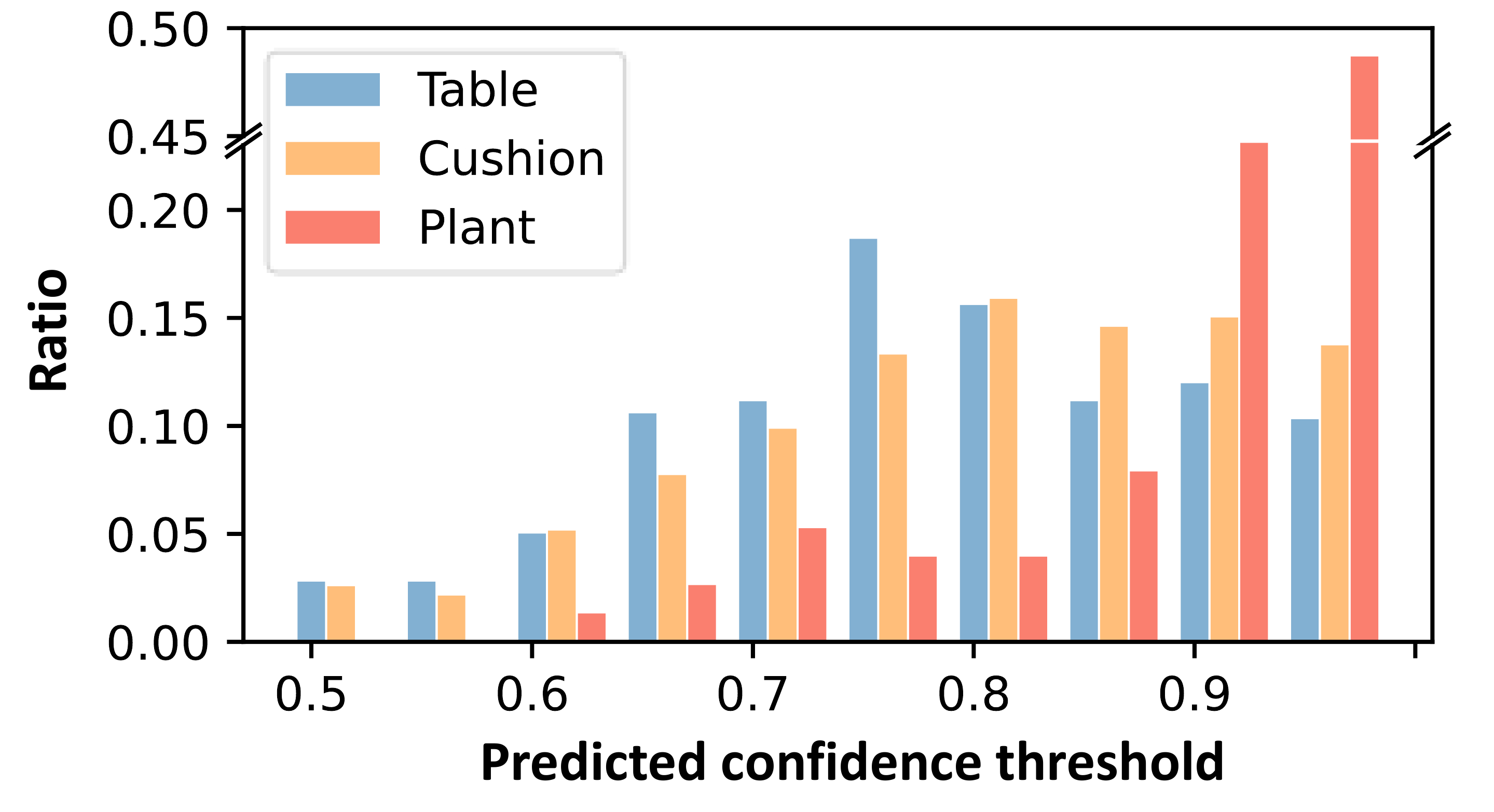}
    % \put(12,30.5){\footnotesize (a)}
    % \put(38,30.5){\footnotesize (b)}
    % \put(65,30.5){\footnotesize (c)}
    % \put(87.5,30.5){\footnotesize (d)}
\end{overpic}
\caption{
An comparison of predicted threshold distribution between different categories by our category-aware policy. We report the ratio of the each predicted threshold. }
\vspace{-0.4cm}
\label{fig:category_aware}
\end{figure}
% (a). \textit{3D Points Fusion.} Based on the back-projection points, we perform 3D points fusion to online organize the 3D points. (b). \textit{Exploration Policy and verification Policy} takes both 2D map (from semantic mapper) and 3D map to predict a discrete direction. \textit{V} takes 3D map and predict a confidence threshold with a label consistency mechanism to determine a final target goal. (c) Finally, a \textit{ local planner} outputs a moving action to reach the target goal/direction. 

\begin{figure}[t]
\centering
\begin{overpic}% 
% \vspace{-0.3cm}
[width=0.9\linewidth]
{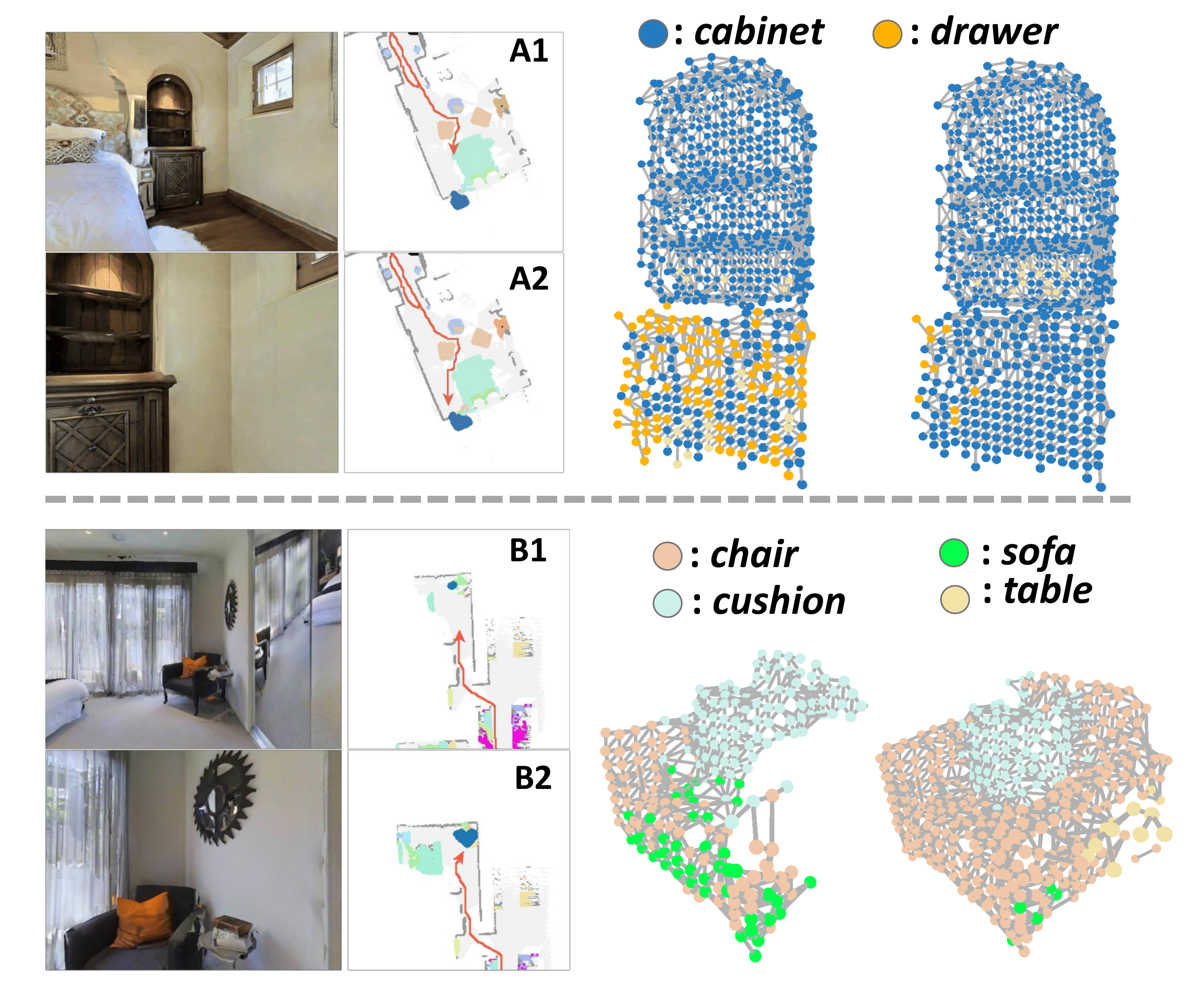}
    % \put(12,30.5){\footnotesize (a)}
    % \put(38,30.5){\footnotesize (b)}
    % \put(65,30.5){\footnotesize (c)}
    % \put(87.5,30.5){\footnotesize (d)}
\end{overpic}
\caption{
Visualization of the results of online 3D point fusion.
}
\vspace{-0.6cm}
\label{fig:result_consistency}
% \end{overpic}
\end{figure}

\noindent\textbf{Comparison on Identification Policy.} Another critical challenge in OjectNav is how to properly identify an instance of target object category. Therefore, We evaluate our identification policy on MP3D along with other identifying strategies, including a 2D frame-based policy adopted in~\cite{chaplot2020object} and 3D point-based methods proposed by our approach.
The results are shown in Table~\ref{tab:verification}. We observe a performance improvement (rows 1 - 2) by simply leveraging 3D point-based construction and fusion algorithm. It can demonstrate that the multi-view observations provide more accurate semantic prediction, which effectively reduces false positive prediction (see examples in Figure~\ref{fig:result_consistency}). Moreover, our category-aware identification policy, through predicting dynamic threshold, demonstrates an even better performance. 

To further investigate the effect of our identification policy, We conduct a break down study in Figure~\ref{fig:category_aware} by plotting the distribution of predicted semantic confidence thresholds.
Specifically, we plot the distribution of three different categories (\textit{table}, \textit{cushion}, \textit{plant}). For a relatively easy-to-recognize category, such as \textit{table} with $52.6$\% success rate (SR), our policy predict a broad threshold distribution. However, for more challenging categories, such as \textit{cushion} ($36.9$\% SR) and \textit{plant} ($16.1$\% SR), the policy tends to be more conservative through setting a higher threshold. The results demonstrate the category-aware characteristic of our identification policy which adapts well to different difficulty levels across categories.

%Note that it can be very tricky to set hard thresholds for different scenes and object category groups, in contrast our predicted threshold can maintain robustness for various settings.

% Meanwhile, the deterministic policy are highly brittle due to the hard confidence threshold.

%a qualitative visualization can be found Figure.\ref{fig:result}. Here, our method online update the semantic prediction and successfully dismiss the wrong target goal. For more episode qualitative results, please refer to appendix or supplementary. 

\noindent\textbf{Ablation Study.} We also perform an ablation study to verify the effectiveness of different components of our method. The results are demonstrated in Table~\ref{tab:ablation}. 
The cooperation of the 2D top-down map and 3D points (row 4) shows significant improvement by incorporating extensive scene perception (in 2D) and fine-grained object perception (in 3D).
Moreover, rows (3-4) and (4-5) proved the effectiveness of leveraging consistency information and the identification policy, respectively.

\begin{table}
\vspace{-10pt}
\caption{Ablation study of main components in our method. The pos. indicates the semantic predictions $p_s$, KL indicates the spatial semantic consistency $p_c$ and the I. policy indicates the usage of the proposed identification policy.}
\scalebox{0.95}{
\setlength{\tabcolsep}{0.5mm}{
\begin{tabular}{cccc|ccc}
\hline
\multirow{2}{*}{2D map} & \multicolumn{2}{c}{3D points} & \multirow{2}{*}{I. Policy} & \multirow{2}{*}{SPL(\%)} & \multirow{2}{*}{Succ.(\%)} & \multirow{2}{*}{DTS(m)} \\
& Pos. & KL & &\\
\hline \hline
    \checkmark &            &            & 
        &  11.2  &  29.6  &  6.213   \\
               & \checkmark & \checkmark & 
        &  13.0  &  32.3  &  5.769  \\
    \checkmark & \checkmark &            & 
        &  13.7  &  33.8  &  5.620  \\
    \checkmark & \checkmark & \checkmark & 
        &  13.9  &  33.5  &  4.931  \\
\hline
    \checkmark & \checkmark & \checkmark & \checkmark
        &  \textbf{14.6}  &  \textbf{34.0}  &  \textbf{4.749}  \\ 
\hline
\end{tabular}}}
\vspace{-0.2cm}
\label{tab:ablation}
\end{table}

% \vspace{5pt}
\noindent\textbf{Analysis of Computational Cost.} Our framework is extremely memory efficient, which requires about $0.5 GB$ for one scene, and can perform online construction and semantic fusion at a frame rate of 15 FPS. Moreover, our method requires only 48 GPU hours to train a 3D-aware agent on MP3D dataset to achieve the SOTA performance among all modular-based methods. This is significantly faster (30x) than other existing reinforcement learning based methods~\cite{ chaplot2020object,ye2021auxiliary}, and is comparable to supervised learning modular-based methods ~\cite{ramakrishnan2022poni}

% \subsection{Qualitative Experiment}

% \input{iclr2023/table/ablation.tex}

% \subsection{Ablation Study}

\vspace{0.1cm}
\section{Conclusion}
\label{sec:conclusion}

In this work, we present a 3D-aware framework for object goal navigation. Our method is based on a 3D point-based construction algorithm to observe the 3D scenes and simultaneously perform exploration and identification polices to navigate the agent. Our method achieve SOTA performance among all modular-based methods, while requiring less training time. In the future, we would like to exploit this 3D-aware framework in other embodied AI tasks, \textit{e.g.} mobile manipulation, robotic nurses.

\noindent\textbf{Acknowledgements.} We thank anonymous reviewers for their valuable suggestions. This work was supported by National Key Research and Development Program of China (2018AAA0102200), NSFC (62132021), and Beijing Academy of Artificial Intelligence (BAAI).

{\small
\bibliographystyle{ieee_fullname}
\bibliography{goalnav}
}

\appendix

% \section{Introduction}
\clearpage
% \Large{\textbf{Appendix}}
\noindent{\fontsize{20}{24}\selectfont\textbf{Appendix}}
\\
\\
We provide additional information about our method, experiment
settings and supporting qualitative visualizations. Below is a summary of the sections in the supplementary:
\begin{itemize}
	\item Section~\ref{sec:points} reports the details of 3D points construction algorithm.
	\item Section~\ref{sec:pipeline} reports the details of simultaneously running exploration and identification policies.
	\item Section~\ref{sec:cc} reports the computational cost of the proposed method.
	\item Section~\ref{sec:dataset} reports the details of datasets used in our experiments.
        \item Section~\ref{sec:add_exp} reports additional experiments.
	\item Section~\ref{sec:detail_vis} shows detailed progressive results of the proposed method on Matterport3D. 
	% \item Section~\ref{sec:moreresults} shows more progressive results of online semantic segmentation on ScanNet\cite{dai2017scannet} dataset.
% 	\item Source code of our network implementation.
% 	\item A live demo video demonstrates the online semantic segmentation performance on a given RGB-D sequence.
\end{itemize}

\section{3D Point Fusion Implementation Details}
\label{sec:points}
\begin{figure}[t]
\centering
\begin{overpic}
[width=1.0\linewidth]
%[width=\linewidth,grid,tics=10]
{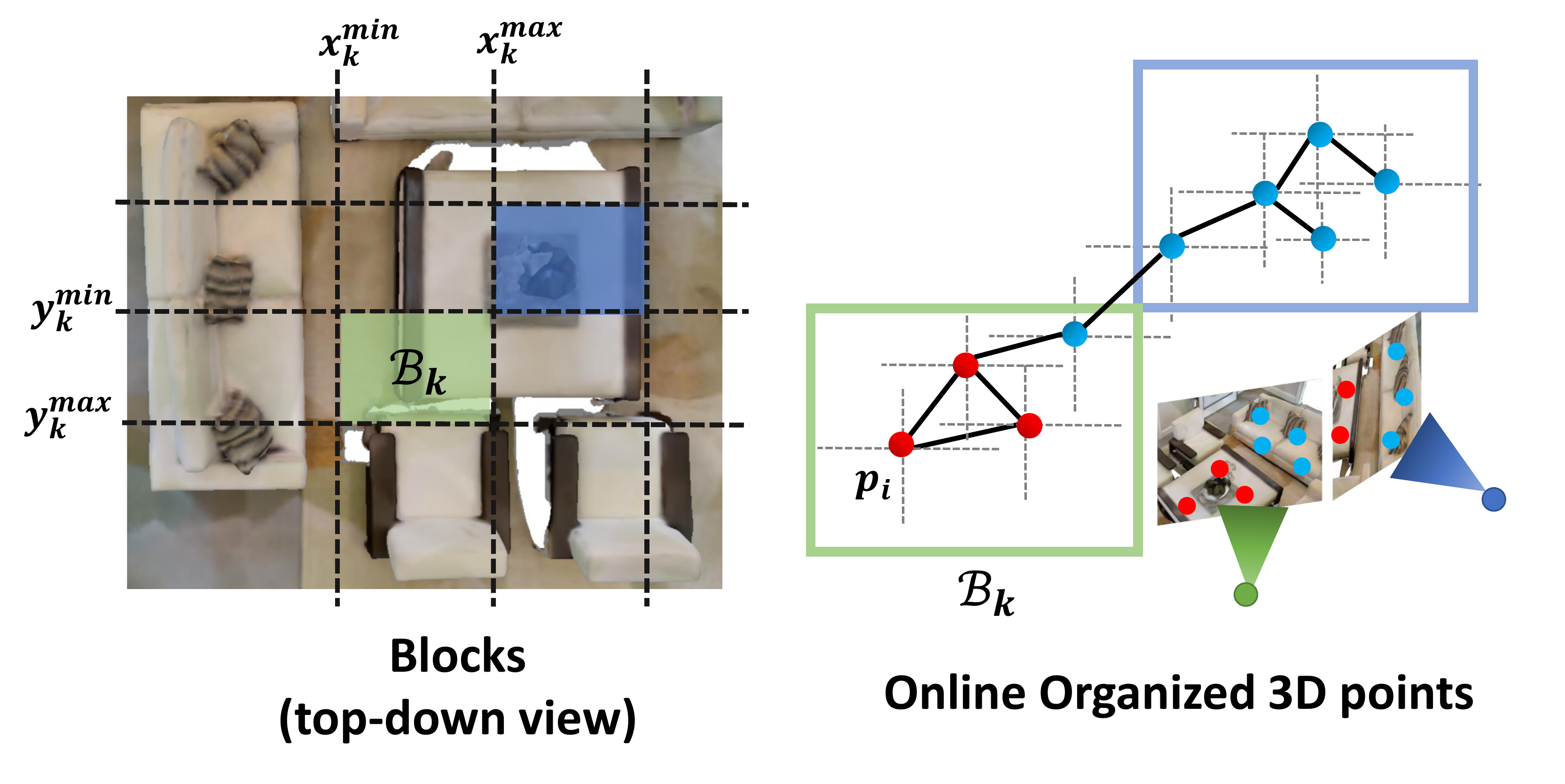}
    % \put(12,30.5){\footnotesize (a)}
    % \put(38,30.5){\footnotesize (b)}
    % \put(65,30.5){\footnotesize (c)}
    % \put(87.5,30.5){\footnotesize (d)}
\end{overpic}
\caption{
Illustration of online 3D point fusion. \textbf{(Left)} dynamically allocated blocks $\B$ based on coordinate intervals . \textbf{(Right)} The points $p$ are organized by blocks $\B$ and per-point octrees $\bO$, which can be used to query neighborhood points of any given point.
}
% \vspace{-0.4cm}
\label{fig:map_suppl}
\end{figure}

We introduce the 3D point construction algorithm~\cite{zhang2020fusion} utilized in our paper (Figure~\ref{fig:map_suppl}). The inputs of the construction algorithm are a sequence of posed color image $I_c^{(t)}$ and depth images $I_d^{(t)}$ at time step $t$. First, we can obtain the 3D points $p$ via back-projection.
Then, we dynamically allocate 3D blocks $\{\B_k\}$, which are composed of the occupied 3D points. 
To be specific, we divide the 3D world space into a set of adjacent 3D blocks $\{B_k\}$, where each block $B_k$ is defined by the boundary of constant length $\tau_b$ along the X, Y and Z axes, \textit{e.g.}, $[X_{min}(B_k), X_{max}(B_k)]$. Two adjacent blocks $B_k, B_j$ along the X axis meet the requirement, $X_{min}(B_k) = X_{max}(B_j)$ or $X_{max}(B_k) = X_{min}(B_j)$. The same requirement holds for Y and Z axes.
Given the scene point cloud $P^{(t-1)}$ at time step $t-1$, we allocate all the 3D points into each of the 3D blocks $\{B_k\}$, hence a block-wise point retrieval can be easily achieved. 

After constructing the blocks, we can achieve efficient point searching and neighborhood retrieval for any given 3D point $p$. However, the points within blocks are still unstructured. 
To obtain the fine-grained relationship of points, we further build a one-level octree $\bO_i$ for each point $p_i \in P$. 
Specifically, for each 3D point back-projected from the instant sensor reading, we perform its nearest neighbor search only among its occupied 3D block and adjacent blocks.
Then, we connect the point with the nearest points in the eight quadrants of the Cartesian coordinate system.
Now, given any point, we can search the nearest points in eight directions and expand the search region as large as we want. In our implementation, we randomly sample 512 points for each frame. And we only connect the points with distance range in $[4cm, 15cm]$. 

\section{Pipeline Implementation Details}
\label{sec:pipeline}
In Algorithm~\ref{algo:pipeline}, we describe the the details of simultaneously running exploration and identification policies. Here, we reuse the notions in the main paper.

\IncMargin{0.5em}
\begin{algorithm}[t!]\scriptsize
	\caption{Simultaneous Exploration and Identification for 3D-Aware ObjectNav}
	\label{algo:pipeline}
	\SetCommentSty{textsf}
	\SetKwInOut{AlgoInput}{Input}
	\SetKwInOut{AlgoOutput}{Output}
	\SetKwFunction{PointFusion}{PointFusion}
	\SetKwFunction{SemanticPredictor}{SemanticPredictor}
	\SetKwFunction{BackProject}{BackProject}
	\SetKwFunction{UpdateConsistency}{UpdateConsistency}
	\SetKwFunction{ExplorationPolicy}{ExplorationPolicy}
	\SetKwFunction{Project}{Project}
	\SetKwFunction{IdentificationPolicy}{IdentificationPolicy}
	\SetKwFunction{ConsistencyCheck}{ConsistencyCheck}
	\SetKwFunction{Location}{Location}
	\SetKwFunction{Planning}{Planning}
	\Indm
	\Indp
	\AlgoInput{Existed reconstructed points $P_{l,s,c}^{(ts)}$, RGB image $I^{(t)}_c$, Depth image $I^{(t)}_d$, Agent Pose $o^{(t)}_{pose}$, Target category index $o_{ID}$ at time step $t$.}
	\AlgoOutput{ Agent action $a^{(t)}$.}
	\tcp{Output a low-level action to drive the agent}
	\tcp{$a^{(t)} \in$ \{\texttt{move\_forward}, \texttt{turn\_left}, \texttt{turn\_right} and \texttt{stop}\}}
	$P_{l,s}^{(t)}\leftarrow$\BackProject{$I_d^{(t)}$,\SemanticPredictor{$I_c^{(t)}$}}\;
	\tcp{Obtain the location and semantic prediction of observed points}
	$P_{l,s}^{(ts)}\leftarrow$\PointFusion{$P_{l,s}^{(ts)}$, $P_{l,s}^{(t)}$}\;
	$P_{c}^{(ts)}\leftarrow$\UpdateConsistency{$P_{l,s}^{(ts)}$}\;
    \tcp{Fuse the new captured points and update the consistency}
	$g_e^{(t)}\leftarrow$\ExplorationPolicy{$P_{l,s,c}^{(ts)}$,\Project{$P_{l,s,c}^{(ts)}$}}\;
    $\tau^{(t)}\leftarrow$\IdentificationPolicy{$P_{l,s,c}^{(ts)}$}\;
    \If{$p\in P_{l,s,c}$ satisfy $p_s > \tau^{(t)}$ and \ConsistencyCheck{$p$,$o_{ID}$,$P_{l,s,c}$}}{$g_f^{(t)}\leftarrow$\Location{$p$}\;}
    \tcp{Simultaneously run the exploration and identification polices}
    \If{$g_f^{(t)}$ exists}{$a^{(t)}\leftarrow$\Planning{$g_f^{(t)}$}\;}
    \Else{ $a^{(t)}\leftarrow$\Planning{$g_e^{(t)}$}\;}
    \Return $a^{(t)}$\;
    % \Else{$a^{(t)}\leftarrow$\Planning{$g_e^{(t)}$}\;\Return $a^{(t)}$\;}
    %
% 	\If{$n$ satisfy $\xmin(n)<x_{p}<\xmax(n)$ exists}{$n\leftarrow$\AddIntoNode{$n,x_{p}$}\;\Return $\gtx,n$}
% 	\tcp{create a node $n$ contains $p$}
% 	\tcp{can pre-create neighbor nodes to reduce creation costs}
% 	$n\leftarrow$\CreateNewNode{$x_{p}$}\;
% 	$interval\leftarrow$\GetInterval{$x_{p},d$}\;
% 	$\xmin(n)\leftarrow interval_{min}$\;
% 	$\xmax(n)\leftarrow interval_{max}$\;
% 	\tcp{Find a nearest node in $R_x$ for n}
% 	$dist, node\leftarrow$\NearestNode{$n,\gtx$}\;
% 	\SetAsChild{$node,n$}\;
	% \If{$dist<d$}
	% {
	% \If{$\xmin(n)<min(node)$}{$\xmax(n)\leftarrow \xmin(node)$}
	% \Else{$\xmin(n)\leftarrow \xmax(node)$}
	% }
% 	\Return $n,\gtx$\;
\end{algorithm}
\DecMargin{0.5em}

\subsection{Policy Implementation Details}

\begin{figure}[t]
\centering
\includegraphics[width=\linewidth]{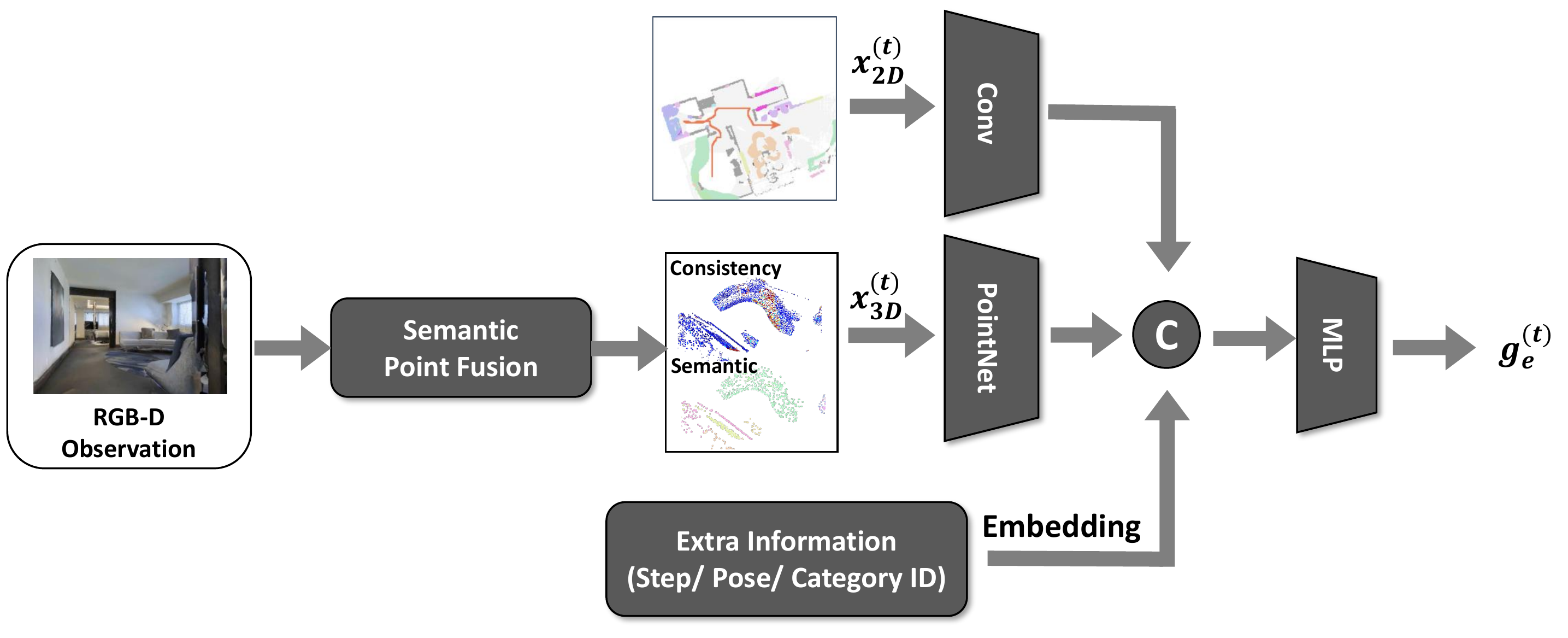}
\caption{Network architecture of our exploration policy}
\label{fig:exp_net}
\end{figure}

Our \textbf{corner-guided exploration policy} takes the 3D observation $x_3D^{(t)}$, 2D observation $x_2D^{(t)}$ and extra information as inputs. The extra information comprises the agent's pose, the number of steps, and the target category ID. The proposed exploration policy predicts a discrete corner goal $g_e^{(t)}$ to navigate the robot (Figure~\ref{fig:exp_net}). Specifically, the policy uses a PointNet~\cite{Qi2017PointNetDL} to encode the 3D points information (position $p_l^{(t)}$, semantics $p_s^{(t)}$, and consistency $p_c^{(t)}$) to obtain a global feature (256D). The 2D top-down map will be passed to a fully convolutional network~\cite{long2015fully} and flattened to a feature vector (256D). And the extra information is embedded into a feature vector (24D). Note that the processing of the 2D top-down map and extra information has also been reported in other existing methods~\cite{chaplot2020object,chaplot2020learning}. Then, the three feature vectors are concatenated and sent to linear networks, which will output the final target corner goal $g_e^{(t)}$. The \textbf{category-aware identification policy} takes 3D observation $x_3D^{(t)}$ and extra information as inputs, and uses the same  3D observation and extra information branches as exploration policy. The identification policy outputs the threshold $\tau^{(t)}$ for target goal selection (See algorithm~\ref{algo:pipeline}).

\begin{figure}[t]
\centering
\includegraphics[width=\linewidth]{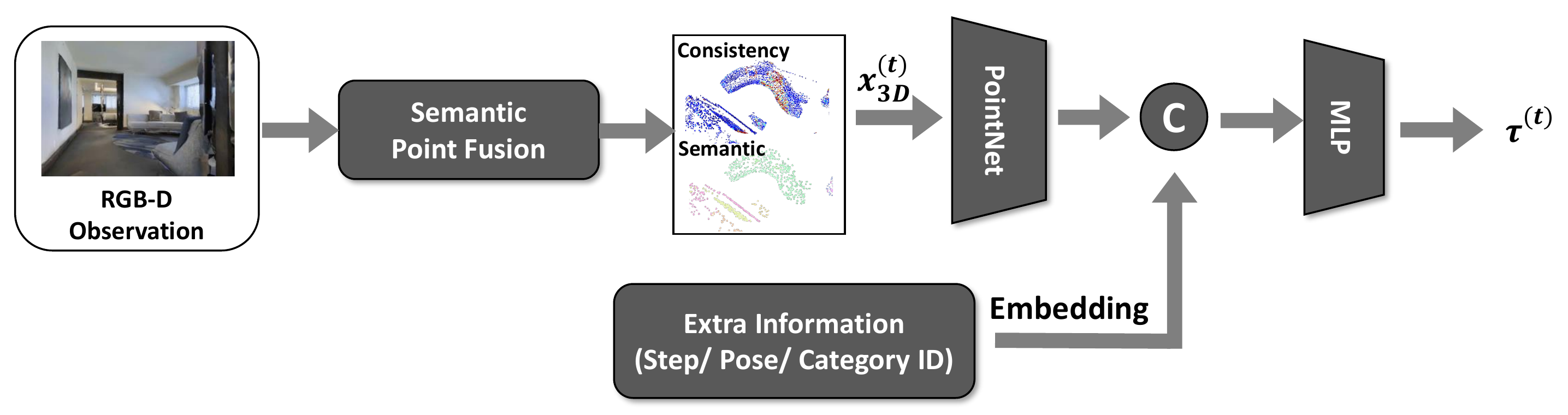}
\caption{Network architecture of our verification policy}
\label{fig:ver_net}
\end{figure}

\section{Computational Cost}
\label{sec:cc}
Due to the fact that there are always significant overlaps between consecutive frames, when we perform point fusion, we can reuse most of the constructed 3D blocks ($\sim60\%$). Our algorithm for constructing the 3D scene representation runs at $15$ FPS. The memory requirement of one scene can range from $200$MB to $500$MB during navigation.

We have implemented our core algorithm in python, PyTorch and PyCUDA. Both the point construction and policies run on a workstation with an Intel\textsuperscript{\textregistered} Xeon\textsuperscript{\textregistered} Gold 6240 CPU CPU @ 3.50GHz × 12 with 64GB RAM and an Nvidia V100 GPU with 32GB memory.
\section{Dataset}
\label{sec:dataset}
Here, we provide further details of the datasets where we validate our method for reference.

\noindent\textbf{Matterport 3D (MP3D)}~\cite{chang2017matterport3d} MP3D offers photorealistic reconstructions of building-scale scenes. Following the setting in Habitat Challenge 2021~\cite{batra2020objectnav}, we consider 21 object categories: \textit{chair}, \textit{table}, \textit{picture}, \textit{cabinet}, \textit{cushion}, \textit{sofa}, \textit{bed}, \textit{chest of drawers}, \textit{plant}, \textit{sink}, \textit{toilet}, \textit{stool}, \textit{towel}, \textit{tv monitor}, \textit{shower}, \textit{bathtub}, \textit{counter}, \textit{fireplace}, \textit{gym
equipment}, \textit{seating} and \textit{clothes}. We split the dataset into 61 train / 11 val scenes, containing 2,632,422 / 2,195 episodes, respectively.

\noindent\textbf{MP3D-L2M.} In L2M\cite{georgakis2022l2m}, they validate their method on a self-made dataset consisting of 781 episodes from 10 MP3D (val) scenes, which we call MP3D-L2M. It covers 6 object categories: \textit{chair}, \textit{couch(sofa)}, \textit{plant}, \textit{bed}, \textit{toilet} and \textit{tv}. For a fair comparison, we also report our validation results on this MP3D-L2M in main paper table 2.

\section{Additional Experiments}
% \vspace{-3.5mm}
\begin{table}[t]
\centering
\scalebox{0.8}{
\setlength{\tabcolsep}{0.8mm}{
\begin{tabular}{l|cc}
\hline
Method                     & SPL(\%)   & ~Improvement   \\ \hline \hline
(1.a) 4 corner goal heuristics w/o iden. policy   & $13.1$ & -\\
(1.b) 4 corner goal heuristics w/ iden. policy    & \textbf{$13.9$} & $6.1\%\uparrow$\\
\hline
(2.a) learn continuous goal policy w/o iden. policy  & $10.1$ & -  \\
(2.b) learn continuous goal policy w/ iden. policy  & \textbf{$12.7$} &  $25.7\%\uparrow$\\
\hline
(\textbf{Ours}) learn 4 corner goal w/o iden. policy  &13.7 & -\\
(\textbf{Ours}) learn 4 corner goal w/ iden. policy  & \textbf{$14.6$} &  $6.6\%\uparrow$\\ \hline
\end{tabular}
}}
% \vspace{-2mm}
\caption{More ablations on exploration and identification policies.}
\label{tab:exploration}
\end{table}
% \vspace{-5mm}
\label{sec:add_exp}
 \textbf{Ablation study on exploration and identification policies.} As shown in Table~\ref{tab:exploration},  1) coupling our identification policy with exploration heuristics~\cite{luo2022stubborn}; 2) joint learning our policy with 2D map-based exploration policy~\cite{chaplot2020object} till \textbf{full convergence} using double training steps of~\cite{chaplot2020object}. 
We observe that the results of various methods are improved by our identification policy, especially the continuous goal exploration strategy ($25.7\%\uparrow$ on SPL). 
Moreover, the fully trained continuous goal strategy does not outperform our corner-guided method, due to larger action space and therefore a harder RL problem. 

% \vspace{-5.5mm}
\begin{table}[t]
\centering
\scalebox{0.85}{
\setlength{\tabcolsep}{0.85mm}{
\begin{tabular}{l|ccc}
\hline
Noise Setting                              & SPL (\%) & Succ. (\%) & DTS (m) \\ \hline \hline
(1) Ours (noiseless)                 & \textbf{$14.6$}     & \textbf{$34.0$}       & \textbf{$4.74$}    \\
(2) w. Noisy Pose                        & $13.5$     & $33.1$       & $5.51$    \\
(3) w. Gau. Noisy Depth                  & $14.3$     & $33.6$       & $5.36$    \\
(4) w. Rdw. Noisy Depth                  & $13.7$     & $31.7$       & $5.45$    \\
(5) w. Noisy Depth (Gau.) and Noisy Pose & $13.6$     & $31.6$       & $5.50$    \\
(6) w. Noisy Depth (Rdw.) and Noisy Pose & $13.1$     & $30.1$       & $6.02$    \\ 
\hline
(7) PONI (baseline, noiseless)                      & $12.1$     & $31.8$       & $5.10$    \\ 
(8) Stubborn (baseline, noiseless)                 & $13.5$     & $31.2$       & $5.01$    \\ \hline
\end{tabular}
}}
% \vspace{-2mm}
\caption{Results under various noise settings. Gau. indicates Gaussian. Rdw. indicates Redwood. }
\label{tab:noise}
\end{table}

\textbf{Robustness to noises.} We conduct a series of experiments (Table~\ref{tab:noise}) to evaluate our method with noisy pose and various depth noise models. 
Specifically, for the pose noise, we adopt the same simulation methods as in~\cite{savva2019habitat}; for depth noise, we consider Redwood~\cite{choi2015robust} noise model and Gaussian noise model. Under the most challenging noises setting (6), our method has a minor drop of $1.5$ point on SPL, which however still performs as a strong competitor against the noiseless baselines (7, 8).

\begin{figure}[t]
\centering
\begin{overpic}
[width=\linewidth]
{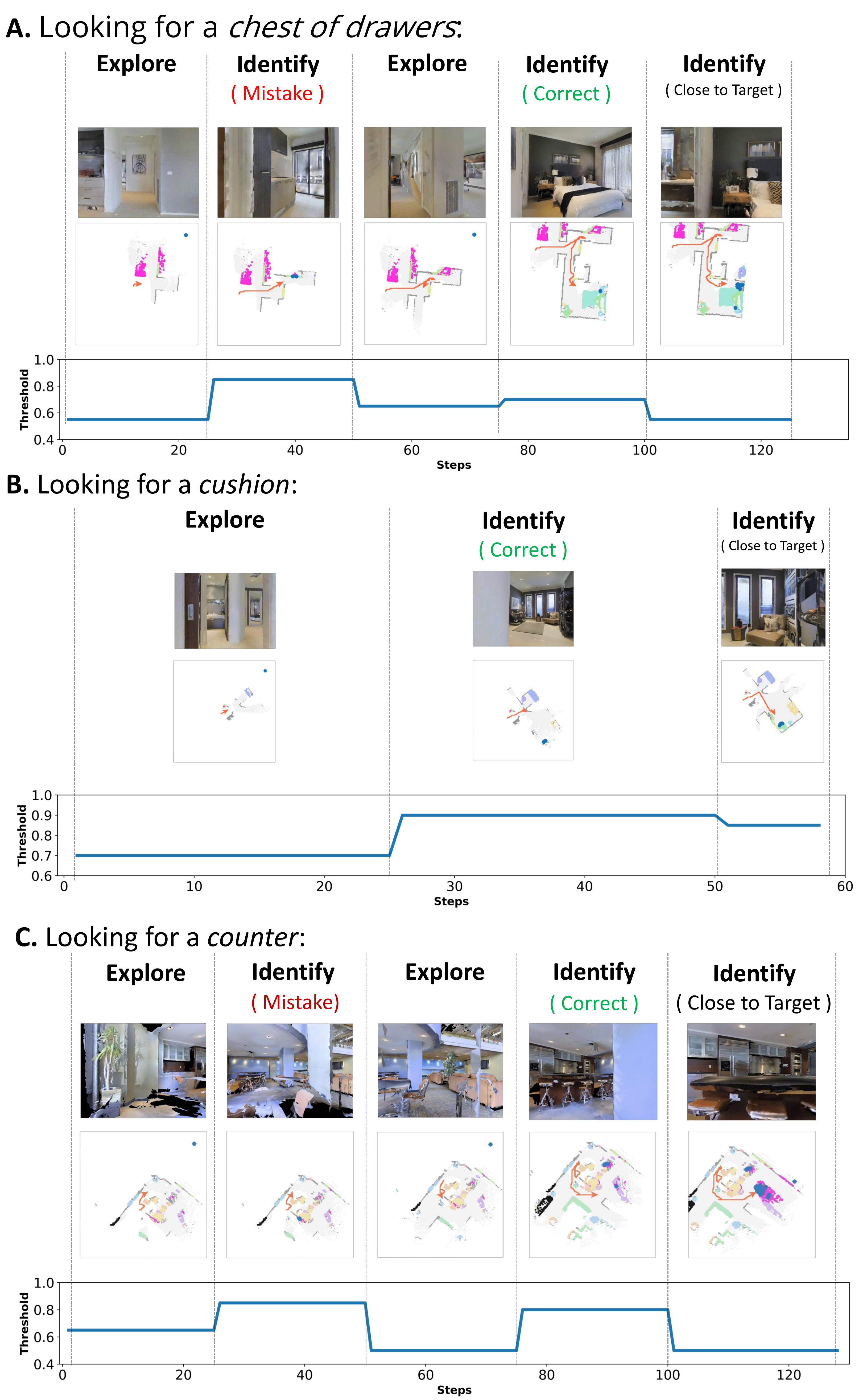}
\end{overpic}
\caption{
Visualization of threshold changes during navigation.
}
\label{fig:vis_thres}
\end{figure}

\textbf{Qualitative examples of the identification policy}. Here we provide three qualitative examples (Figure~\ref{fig:vis_thres}) of our identification policy to navigate to various target objects. Note that, the identification policy is executed every 25 steps for the purpose of acceleration. Accordingly, we delineate the exploration and identification phases based on the primary policy utilized during the 25-step interval.

Based on the experimental results, the predicted threshold is dynamically adjusted during navigation. Specifically, when the agent is under the control of the exploration policy, the identification policy predicts a relatively low threshold to facilitate rapid searching of potential targets. Conversely, when the agent is guided by the identification policy, the threshold is fine-tuned to achieve a trade-off between accuracy and efficiency. In general, the identification policy gives a high threshold to ensure a successful stop. Nevertheless, in the event that the agent is in close proximity to the target object, a low threshold is predicted for a quick stop.

% adjust to a high threshold level. In this way, the agent requires more accurate perception to decide the final stop. 

% We observe that the predicted threshold is relatively low during the exploration phase (A, C), which could increase the probability of finding any potential target object. 
% Once switching to identification (when the agent finds potential targets (B, D)), a higher threshold is given to filter out the wrong semantic predictions. Finally, the threshold is decreased for a quick stop when the agent comes close to the target (E).  

\section{Visualizing ObjectNav episodes}
\label{sec:detail_vis}
\begin{figure*}[t]
\centering
\begin{overpic}
[width=0.85\linewidth]
%[width=\linewidth,grid,tics=10]
{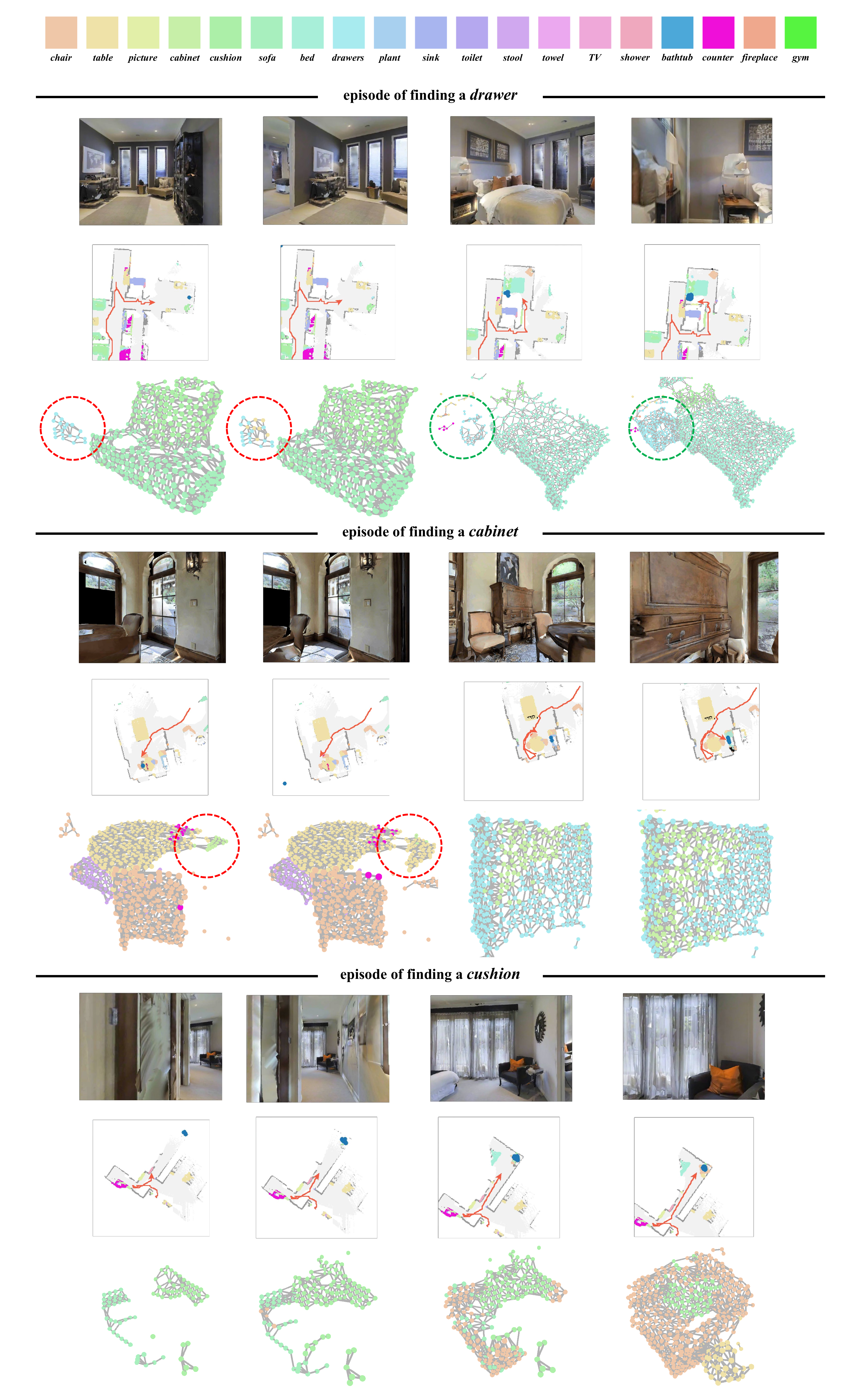}
    % \put(12,30.5){\footnotesize (a)}
    % \put(38,30.5){\footnotesize (b)}
    % \put(65,30.5){\footnotesize (c)}
    % \put(87.5,30.5){\footnotesize (d)}
\end{overpic}
\caption{
EPS Page1. 
}
\label{fig:appendix1}
\end{figure*}

\begin{figure*}[t]
\centering
\begin{overpic}
[width=0.85\linewidth]
%[width=\linewidth,grid,tics=10]
{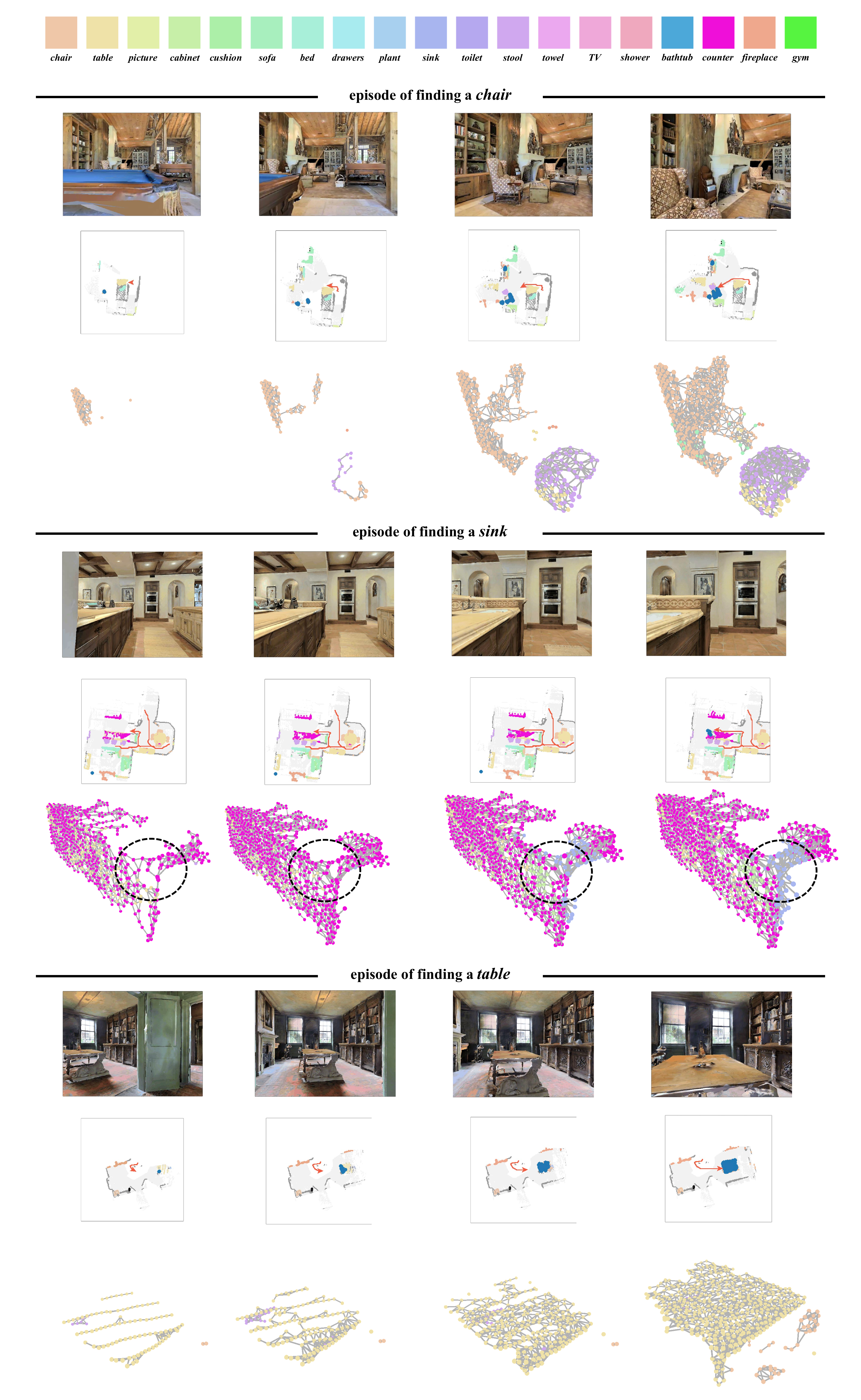}
    % \put(12,30.5){\footnotesize (a)}
    % \put(38,30.5){\footnotesize (b)}
    % \put(65,30.5){\footnotesize (c)}
    % \put(87.5,30.5){\footnotesize (d)}
\end{overpic}
\caption{
EPS Page2. 
}
\label{fig:appendix2}
\end{figure*}

\begin{figure*}
\centering
\begin{overpic}
[width=0.9\linewidth]
%[width=\linewidth,grid,tics=10]
{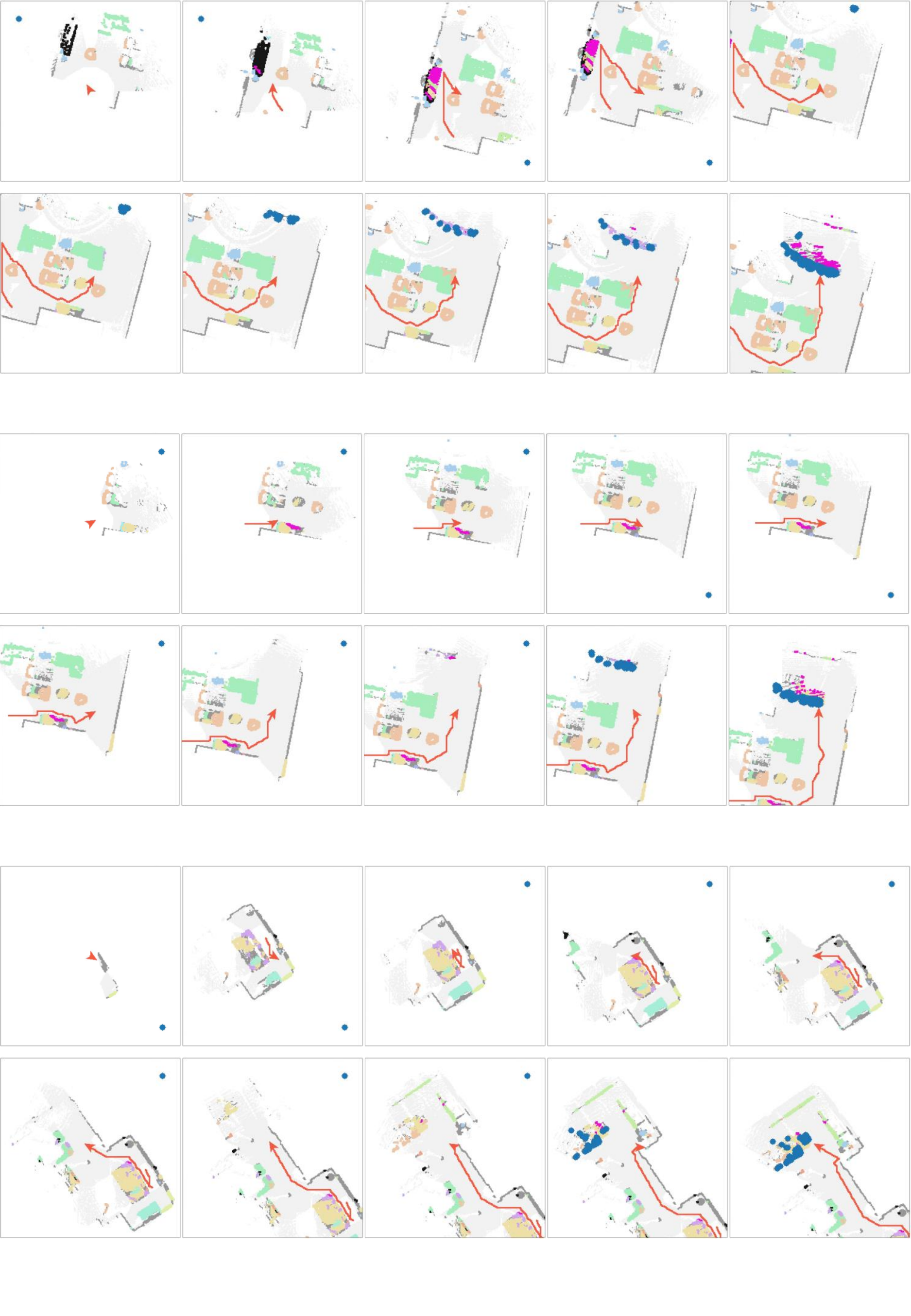}
    % \put(12,30.5){\footnotesize (a)}
    % \put(38,30.5){\footnotesize (b)}
    % \put(65,30.5){\footnotesize (c)}
    % \put(87.5,30.5){\footnotesize (d)}
\end{overpic}
\caption{
Episode results (1/3). 
}
\label{fig:eps_1}
\end{figure*}

\begin{figure*}
\centering
\begin{overpic}
[width=0.9\linewidth]
%[width=\linewidth,grid,tics=10]
{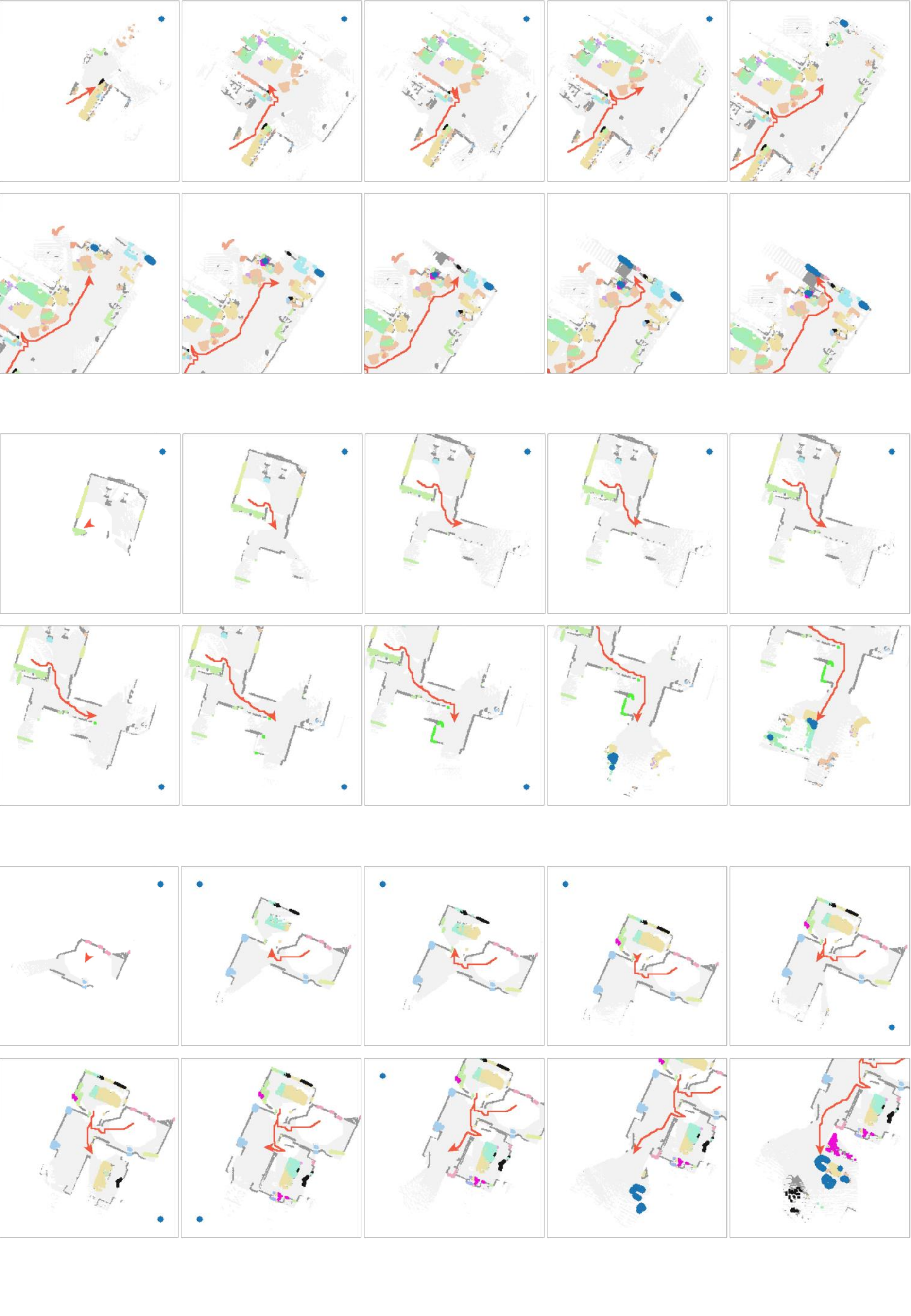}
    % \put(12,30.5){\footnotesize (a)}
    % \put(38,30.5){\footnotesize (b)}
    % \put(65,30.5){\footnotesize (c)}
    % \put(87.5,30.5){\footnotesize (d)}
\end{overpic}
\caption{
Episode results (2/3). 
}
\label{fig:eps_2}
\end{figure*}

\begin{figure*}
\centering
\begin{overpic}
[width=0.9\linewidth]
%[width=\linewidth,grid,tics=10]
{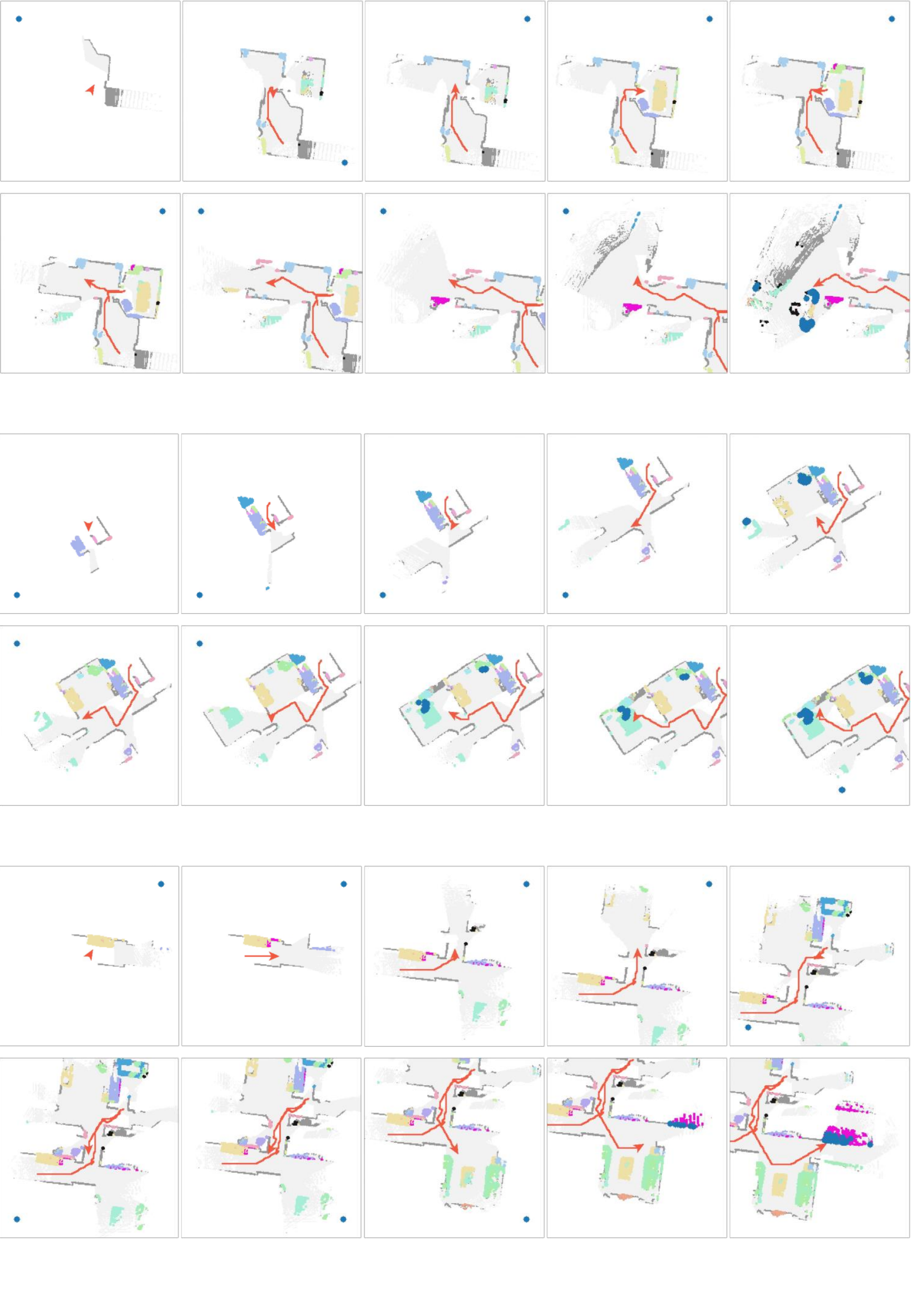}
    % \put(12,30.5){\footnotesize (a)}
    % \put(38,30.5){\footnotesize (b)}
    % \put(65,30.5){\footnotesize (c)}
    % \put(87.5,30.5){\footnotesize (d)}
\end{overpic}
\caption{
Episode results (3/3). 
}
\label{fig:eps_3}
\end{figure*}

Figure~\ref{fig:appendix1} and Figure~\ref{fig:appendix2} illustrate a more detailed visualization of the results obtained through the implementation of our 3D point-based fusion algorithm on the Matterport3D Dataset. It can be observed that during navigation, there is a significant improvement in the semantic prediction and spatial consistency of the points. To further substantiate these findings, Figure~\ref{fig:eps_1} to \ref{fig:eps_3} provide visualizations of additional episodes. For a comprehensive demonstration, the attached video in the supplemental material is recommended.

% Figure~\ref{fig:appendix1} and Figure~\ref{fig:appendix2} present some visual results of our 3D points-based fusion algorithm on Matterpor3D Dataset. We show that the semantic prediction and spatial consistency of points are getting bettering during navigation. For live demo, please refer to the attached video in the supplemental material. 

\end{document}